\documentclass{article}

\PassOptionsToPackage{numbers, compress}{natbib}
\usepackage{natbib}

\usepackage{iclr2026_conference,times}
\iclrfinalcopy

\usepackage[utf8]{inputenc} % allow utf-8 input
\usepackage[T1]{fontenc}    % use 8-bit T1 fonts
\usepackage{hyperref}       % hyperlinks
\usepackage{url}            % simple URL typesetting
\usepackage{booktabs}       % professional-quality tables
\usepackage{amsfonts}       % blackboard math symbols
\usepackage{nicefrac}       % compact symbols for 1/2, etc.
\usepackage{microtype}      % microtypography
\usepackage{xcolor}         % colors
\usepackage{relsize}
\usepackage{longtable}
\usepackage{tabularx}

\usepackage{microtype}
\usepackage{graphicx}
\usepackage{subcaption}
\usepackage{booktabs} % for professional tables
\usepackage{adjustbox}
\usepackage{makecell}
\usepackage{booktabs} 
\usepackage{amsmath}
\newcommand{\minisection}[1]{\noindent{\textbf{#1}.}}

\newcolumntype{T}[1]{>{\ttfamily\RaggedRight\arraybackslash\hspace{0pt}\collectcell\seqsplit}p{#1}<{\endcollectcell}}

\hypersetup{
    colorlinks=true,
    linkcolor=[RGB]{0,0,140},
    citecolor=[RGB]{50,0,90},
    filecolor=black,
    menucolor=black,
    urlcolor=[RGB]{10,0,90},
    breaklinks=true
}

\title{\centering Visualizing Thought: Conceptual Diagrams Enable Robust Planning in LMMs}

\newcommand{\namegap}{\\[4pt]}     % gap between the two name rows
\author{
\begin{tabular}{c}
\namegap
\textbf{Nasim Borazjanizadeh}\textsuperscript{1,2}\hspace{3em}
\textbf{Roei Herzig}\textsuperscript{2}\hspace{3em}
\textbf{Eduard Oks}\textsuperscript{1}\\[4pt]
\textbf{Trevor Darrell}\textsuperscript{2}\hspace{3em}
\textbf{Rogerio Feris}\textsuperscript{3}\hspace{3em}
\textbf{Leonid Karlinsky}\textsuperscript{1}\\[6pt]
\normalfont\textsuperscript{1}Xero Inc.\hspace{2em}
\normalfont\textsuperscript{2}Berkeley AI Research, UC Berkeley\hspace{2em}
\normalfont\textsuperscript{3}MIT--IBM Watson AI Lab
\end{tabular}
}

\usepackage[moderate]{savetrees}

\begin{document}
\maketitle

\vspace{-0.4cm}
\begin{abstract}
\vspace{-0.2cm}
Human reasoning relies on constructing and manipulating mental models—simplified internal representations of situations used to understand and solve problems. Conceptual diagrams (e.g., a sketch drawn to aid reasoning) externalize these mental models, abstracting irrelevant visual details to efficiently capture how entities interact. In contrast, Large Language Models (LLMs) and Large MultiModal Models (LMMs) predominantly reason through text, limiting their effectiveness on complex multi-step tasks. In this paper, we propose Visual Thinking, a generalizable framework that enables LMMs to reason through multiple chains of self-generated conceptual diagrams, significantly enhancing their combinatorial planning capabilities. Our approach requires no human input beyond the natural language description of the task. It integrates textual and diagrammatic reasoning within an optimized Graph-of-Thought inference framework, enhanced by beam search and depth-wise backtracking. Evaluated on multiple challenging PDDL planning domains, our method substantially improves LMM performance (e.g., GPT-4o: $35.5\% \rightarrow 90.2\%$ in Blocksworld) and consistently outperforms text-only search-based inference methods. Additionally, on more difficult planning domains with solution depths up to 40, our approach outperforms the o1-preview reasoning model (e.g., 16 percentage points improvement in Floor Tiles). These results highlight the value of conceptual diagrams as a reasoning medium in LMMs.

\vspace{-0.2cm}
\end{abstract}

% Requires \usepackage{booktabs,makecell,caption}
% Fits on one line in single-column layouts. If tight, reduce \tabcolsep below 2.6pt.

\section{Introduction}\label{sec:intro}
\vspace{-0.1cm}

Natural language is a powerful medium for communication, enabling humans to effectively share knowledge and ideas~\cite{pinker1990natural, clark1996using, tomasello2010origins}. However, language alone is not an optimal medium for reasoning, as it is inherently linear, sequential, and verbose, making it inefficient for representing complex logical and relational structures~\cite{larkin1987diagrammatic, hegarty2004mechanical, tversky2011visualizing}. Prior evidence `human thought' is inherently not verbal, sequential, or linear; rather, it is spatial, parallel, and image-like~\cite{tversky2011visualizing}. Humans construct and utilize internal mental models—simplified analogues of real or hypothetical situations ~\cite{johnson1983mental, gentner2001mental, hegarty2004mechanical, battaglia2013simulation}, and dynamically manipulate them to represent and predict interactions between objects and solve problems. Crucially, mental models are multimodal, integrating both visual and verbal representations to facilitate learning and robust reasoning~\cite{mayer2002multimedia}. Finally, visual representations have always played a central role in human reasoning and communication, from prehistoric cave art, which predates written language~\cite{clottes2008cave}, to modern textbook diagrams, scientific figures, and blackboard sketches.

%Prior evidence~\cite{johnson1983mental, gentner2001mental, hegarty2004mechanical, battaglia2013simulation} indicates that humans do not rely solely on language when reasoning. Instead, they construct and utilize internal mental models—simplified analogues of real or hypothetical situations. Humans dynamically manipulate these mental models through mental simulations, which represent and predict interactions between objects, to solve problems. Crucially, mental models are multimodal, integrating both visual and verbal representations to facilitate learning and robust reasoning~\cite{mayer2002multimedia}. Cognitive science research demonstrates that `human thought' is not inherently verbal, sequential, or linear; rather, it is spatial, parallel, and image-like~\cite{tversky2011visualizing}. Finally, visual representations have always played a central role in human reasoning and communication, from prehistoric cave art—one of the earliest known forms of communication, predating written language~\cite{clottes2008cave}—to contemporary textbook diagrams, scientific figures in articles, and blackboard sketches.

\vspace{-0.13cm}

Conceptual diagrams are simplified visual representations that use basic shapes (e.g., circles, squares, lines) to capture how entities interact while abstracting away irrelevant details~\cite{tversky2011visualizing, larkin1987diagrammatic}. They externalize internal mental models, reducing cognitive load and enabling rapid perceptual inference and clearer reasoning~\cite{hegarty2004mechanical, gentner2001mental}. Unlike photorealistic images, which capture fine-grained details of how objects appear, conceptual diagrams encode the structural and relational information essential for reasoning, using colors, relative positions and sizes, and annotations~\cite{tversky2011visualizing, larkin1987diagrammatic}. For example, a square in a diagram might represent a complex object such as a car, with its color, relative size, and position visually encoding relationships to other entities while omitting irrelevant appearance details. Thus, conceptual diagrams are an effective reasoning medium complementary to language, overcoming language’s limitations in representing relational structure and aligning closely with humans’ multimodal reasoning~\cite{tversky2011visualizing, hegarty2004mechanical, dziri2023faith}.

\vspace{-0.13cm}
Modern large language models (LLMs) and large multimodal models (LMMs)~\cite{OpenAI2023GPT4TR, openai_gpt4o, claude3} have achieved remarkable success on mathematical and scientific benchmarks, including GSM8K~\cite{cobbe2021gsm8k}, MATH~\cite{hendrycks2021measuring}, and GPQA~\cite{zhong2023gpqa}. Despite these advances, their reasoning remains inconsistent, particularly on multi-step compositional reasoning, long-horizon planning, and tasks requiring backtracking or error correction~\cite{dziri2023faith, valmeekam2023planning, creswell2023selection, borazjanizadeh2024navigating}. These limitations stem partly from LLMs’ reliance on language, which is inherently linear and inefficient for representing complex relational structures~\cite{dziri2023faith, valmeekam2023planning, borazjanizadeh2024reliable}. Moreover, the autoregressive architecture of current models enforces sequential next-token prediction, making backtracking challenging~\cite{dziri2023faith}. Thus, enabling LMMs to reason with conceptual diagrams and backtrack within a graph-based inference framework offers a promising approach to overcome these bottlenecks.

\begin{figure}[t] 
    \vspace{-0.78cm}
    \centering
    \includegraphics[width=0.81\columnwidth]{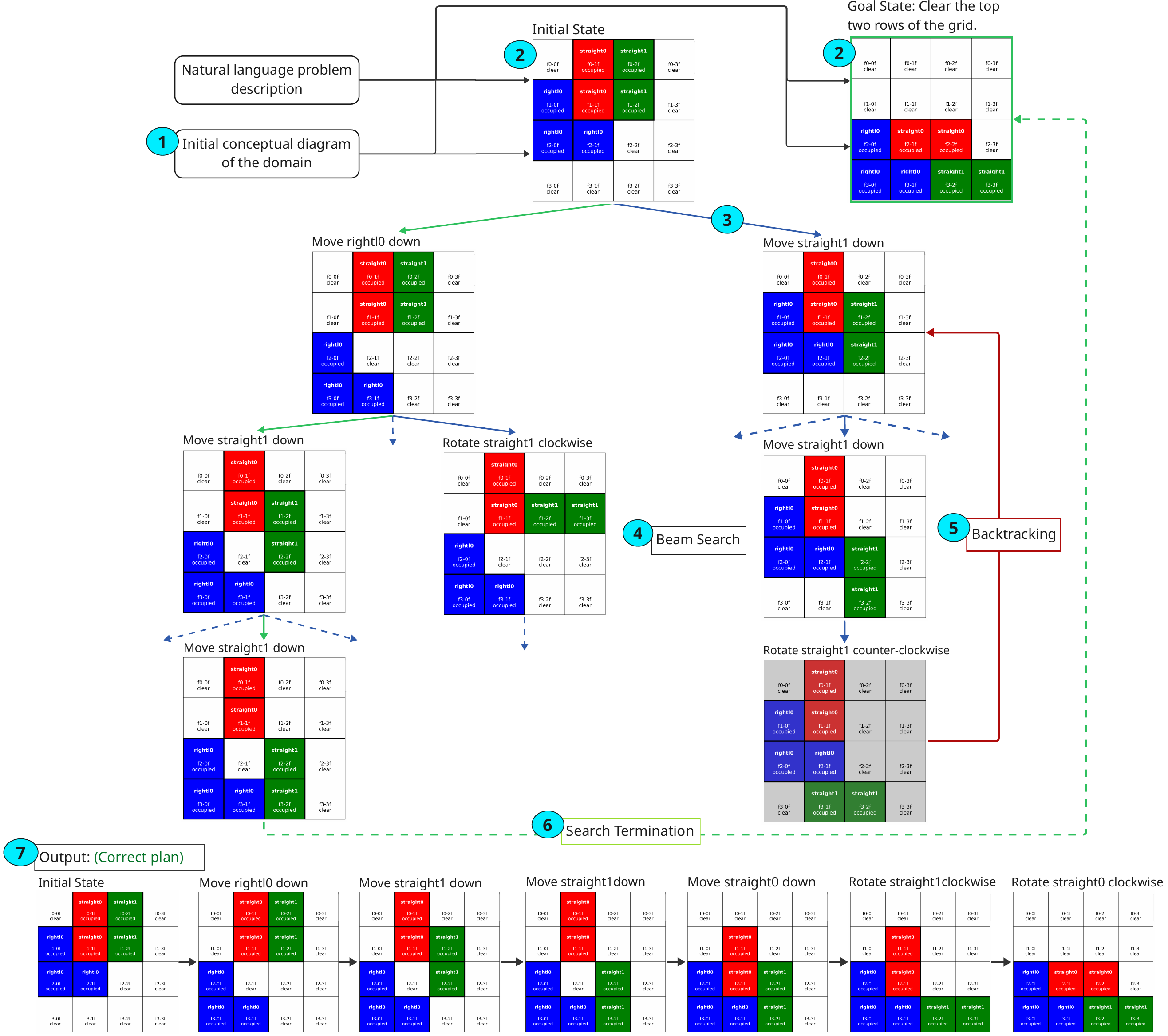}
    \vspace{-0.13cm}
\caption{\small \textbf{Our proposed approach}.  Example diagrams are from the Tetris domain, where tiles are moved on a grid to reach a goal state. (1) The model generates multiple diagram schemas and codes for a random instance; their rendered diagrams are ranked, and the code of the top choice is cached (Fig.~\ref{fig:conceptual_diagram}). (2) Conditioned on this code, diagrams for the initial and goal states are generated (Fig.~\ref{fig:child_gen}). (4) Beam search ranks all candidates at a depth by proximity to the goal, expanding the top $k=4$ (Sec.~\ref{sec:method}). (5) Depth-wise backtracking is applied when all candidate states at a depth fail validation, returning to the deepest available ancestor. (6) The process stops when the goal is reached or a maximum number of steps is visited. (7) The output is the action sequence (plan) plus textual and diagrammatic representations of intermediate states.\textsuperscript{1}}
\vspace{-0.58cm}
    \label{fig:overall}
\end{figure}

\begin{figure}[t] 
    \centering
    \vspace{-0.61cm}
    \includegraphics[width=0.8\columnwidth]{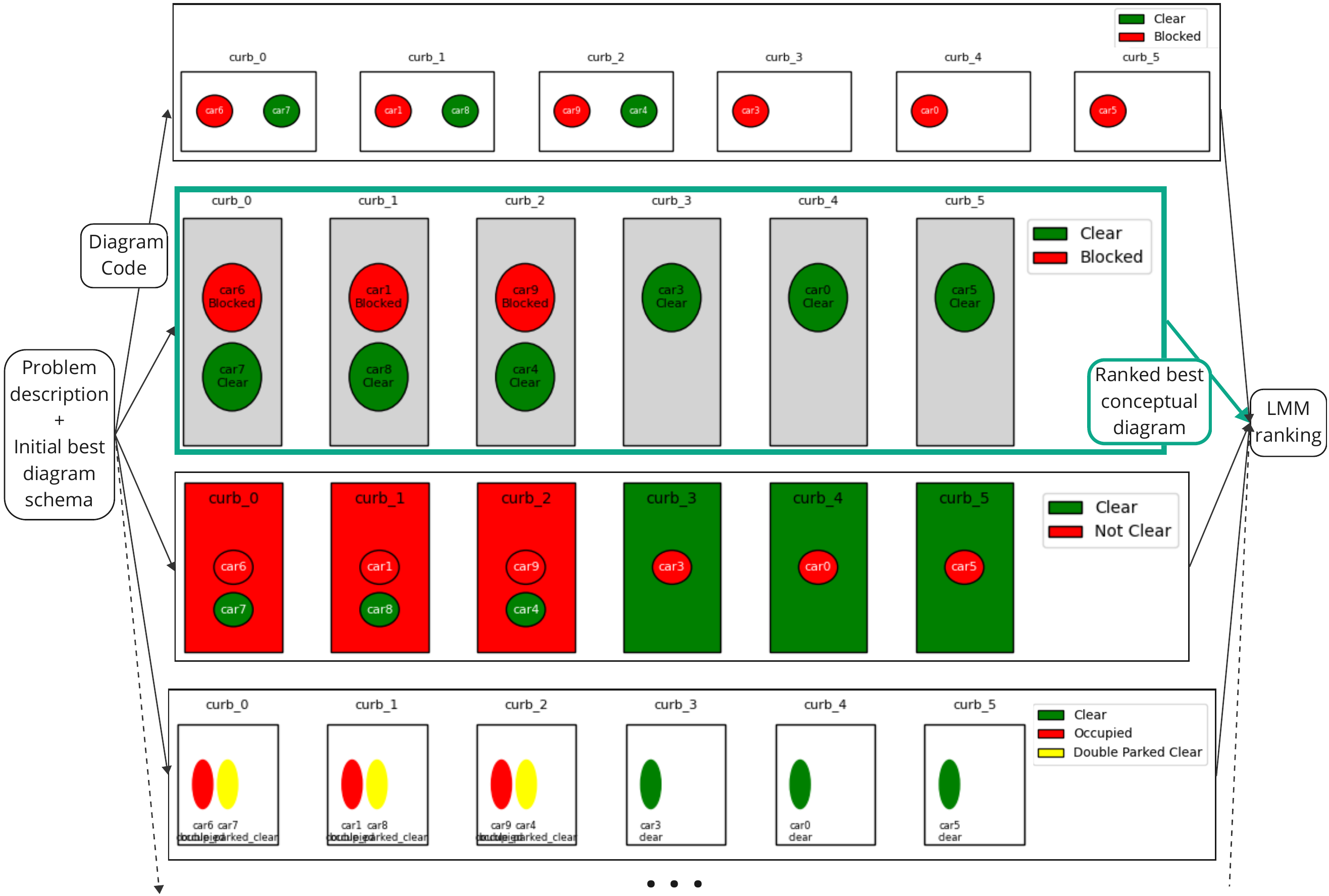}
    \vspace{-0.15cm}
    \caption{\small We generate an initial conceptual diagram for each domain by sampling multiple diagram codes. An LMM ranks diagrams based on intuitive and accurate visualization of relational information. The top-ranked diagram's code serves as reference for generating initial and goal state diagrams for all instances. Example shown is the parking domain, where curbs hold up to two cars, and cars can be movable or blocked (double-parked).}
    \vspace{-0.55cm}
    \label{fig:conceptual_diagram}
\end{figure}

\footnotetext[1]{\relsize{0.6} 
For Figs.~\ref{fig:overall} and \ref{fig:state_chains}, we adjusted the diagram codes only to increase font sizes for better readability. 
Figs.~\ref{fig:conceptual_diagram}-\ref{fig:child_gen} are unmodified generations. Fig.~\ref{fig:overall} actions are simplified; full action strings include previous/current cells occupied.}

\vspace{-0.13cm}

In this work, we propose Visualizing Thought, a framework that enables LMMs to solve combinatorial problems through multiple multimodal chains of self-generated conceptual diagrams and textual reasoning. Our approach requires no domain-specific modifications or manual engineering to solve any combinatorial problem expressible in the Planning Domain Definition Language (PDDL)~\cite{mcdermott1998pddl}, given only a natural language specification of the initial state, goal state, and possible actions. Importantly, our method does not rely on predefined visual templates or geometric priors; instead, it generates conceptual diagrams directly from the textual problem description. 

\vspace{-0.13cm}

Visualizing Thought decomposes inference into a graph of intermediate reasoning steps, where at each node the model selects the next best state. Each node is multimodal, containing both a textual description of the state and a corresponding conceptual diagram (see Fig.~\ref{fig:overall}). At each step, the LMM (i) generates the next state conditioned on the textual and diagrammatic representations of the states in the action path; (ii) produces a diagram schema—a structured set of statements specifying each object’s shape, relative size and location, and status; and (iii) generates Matplotlib code from the schema that renders the state diagram. To ensure inference quality, we incorporate guardrails such as diagram-schema self-reflection checks and local (parent–child) and global (entire path from the initial state) validity checks. To manage the exponential growth of the combinatorial search space~\cite{besta2023graph,yao2023tree}, we integrate beam search to rank validated candidate states at each inference depth and expand only the top \( k\). We also incorporate depth-wise backtracking, which allows the model to revisit earlier validated nodes if all current candidates fail verification. Together, these components enable more efficient exploration of the search space.

\vspace{-0.13cm}

Unlike prior approaches that augment language models with visual representations for compositional reasoning~\cite{wu2024imagine, hu2024visual}—which typically provide an initial visual template for the model to iteratively update—Visualizing Thought relies solely on textual descriptions. The model autonomously generates conceptual diagrams from scratch for every state, without any human-provided visual examples or cues, mirroring how humans use imagination to construct mental models from language. Moreover, instead of producing a single static visualization~\cite{hu2024visual, wang2024whiteboard}, our method generates evolving sequences of intermediate diagrams that  illustrate how the LMM's `model' of the problem evolves with each reasoning step.

\begin{table*}[t]
\centering
\vspace{-0.54cm}
\setlength{\tabcolsep}{6pt} % adjust column padding
\renewcommand{\arraystretch}{01.02}
{\fontsize{9}{11}\selectfont  % or remove entirely for normal size

\begin{tabular}{lccccccc}
\toprule
& \multicolumn{3}{c}{Single inference} & \multicolumn{4}{c}{Search-based inference (base LLM = GPT-4o)} \\
\cmidrule(lr){2-4} \cmidrule(lr){5-8}
\textbf{Domain}
& \makecell{GPT-4o}
& \makecell{o1-\\mini}
& \makecell{o1-\\preview}
& \makecell{GoT}
& \makecell{Optimized\\GoT}
& \makecell{RAP\textsuperscript{(20)}}
& \makecell{Visual Thinking } \\
\midrule
Blocksworld (simple) & 35.5\%$^{*}$ & 56.6\% $^{*}$  & \textbf{97.8\%}$^{*}$ & 50\% & 58\% & 58\% & 90.2\% \\
Blocksworld (hard)   & 0\%          & 0.9\% $^{*}$ & 23.65\%$^{*}$        & 8\%  & 48\%   & 4\%  & \textbf{78\%} \\
Floor Tiles          & 0\%          & 6\% & 20\%                & 0\%  & 4\%  & --   & \textbf{36\%} \\
Parking              & 2\%          & 8\% & 40\%                & 14\% & 28\% & --   & \textbf{52\%} \\
Tetris               & 0\%          & 2\% & 26\%            & 0\%  & 12\% & --   & \textbf{38\%} \\
Elevator             & 0\%          & 2\% & 36\%               & 2\%  & 10\% & --   & \textbf{48\%} \\
Barman               & 0\%          & 0\%   & 10\%                & 0\%  & 4\%  & --   & \textbf{30\%} \\
\bottomrule
\end{tabular}}

\vspace{-0.2cm}
\caption{\small Accuracy results across all evaluated domains. Each baseline was evaluated on 50 problem instances per domain, except for GPT-4o + Visual Thinking on Blocksworld (simple), where we evaluated 500 instances (the full PlanBench~\cite{valmeekam2023planbench}) and achieved 90.2\% (451/500). Baseline results marked with $^{*}$ are taken from \cite{valmeekam2024planning}. RAP reported 51\% accuracy on Blocksworld (simple) using Llama 2~\cite{hao2023reasoning}.}
\vspace{-0.67cm}
\label{tab:all_results}
\end{table*}

\vspace{-0.13cm}

Evaluations across multiple challenging PDDL planning domains demonstrate that our method substantially enhances LLMs’ combinatorial reasoning capabilities (Tab.~\ref{tab:all_results}). On the widely studied Blocksworld domain, from PlanBench~\cite{valmeekam2023planbench}, our approach delivers performance gains of 43, 64, and 55 percentage points using Claude 3.5 Sonnet~\cite{sonnet}, Llama 4 Maverick~\cite{Llama4}, and GPT‑4o, respectively (e.g., GPT-4o’s accuracy rises from 35.5\% to 90.2\%). Importantly, we contribute \textit{a new, more difficult planning benchmark} with five additional planning domains—Floor Tiles, Parking, Tetris, Elevator, and Barman—with solution depths designed up to 40. On this benchmark, our method succeeds where base models consistently fail (e.g., 36\% vs. 0\% in Floor Tiles). Furthermore, Visual Thinking (using GPT-4o) outperforms the reasoning model, o1-preview~\cite{openai_o1preview}, across all new domains (e.g., 10\% vs. 30\% in Barman). Finally, compared to strong search-based inference methods such as Graph-of-Thought~\citep{besta2023graph} and RAP~\cite{hao2023reasoning} (a Monte Carlo Tree Search framework that uses an LLM to build world models and generate plans), our method improves accuracy by at least 22 percentage points while also reducing inference cost by over 30\% and latency by more than 25\%. 

\vspace{-0.13cm}

Crucially, our ablation study on the Blocksworld (simple) domain shows that it is the representation of relational information in conceptual diagrams, not merely the encoded content, that drives these gains. Replacing rendered diagrams with their underlying Matplotlib code, which contains the same spatial and relational data, caused accuracy to collapse from 90.2\% to 24\%, below the GPT-4o single-inference baseline (35.5\%) (Table~\ref{tab:ablation}). This sharp decline shows that the compact, parallel, multi-dimensional (2D, color-encoded) representation of object interdependencies in diagrams, rather than their sequential form in text or their syntactically cluttered code representation, is what enables more effective reasoning.

\vspace{-0.13cm}

To summarize, our contributions include: 1) a cognitively inspired reasoning framework, Visualizing Thought, that enables LMMs to reason with conceptual diagrams autonomously generated from textual descriptions, with no manual engineering or visual templates required for new domains, within a structured graph-based inference process; 2) empirical evidence that representation of information, not just the content, is critical for reasoning, as replacing rendered diagrams with their code containing the same data causes performance to collapse; and 3) extensive evaluations on PlanBench and a new benchmark of five long-horizon planning domains, where our method consistently outperforms strong search-based baselines and reasoning models (on new domains), demonstrating that conceptual diagrams enable solving problems beyond the reach of purely textual (single-inference or search-based) approaches.

\vspace{-0.2cm}
\section{Related Work}
\vspace{-0.2cm}
\minisection{Multimodal Chain-of-Thought for Reasoning} Several recent works have explored integrating visual representations into the reasoning processes of LLMs and LMMs. \citet{hu2024visual} equips LMMs with drawing tools to graph equations or mark photorealistic images, but primarily focuses on single-step or shallow problems. Similarly, \citet{wang2024whiteboard}  generates visual aids for spatial reasoning, providing a single refined visualization per problem. In both works, the generated visualizations are typically an approximate or augmentation of high-fidelity illustrations rather than conceptual diagrams drawn using a model-defined mapping of complex entities to simple shapes and colors. Concurrent work, \citet{wu2024imagine}, also generates visual and textual intermediate states but requires conditioning the model on a human-provided initial visual representation that supplements the textual description of the problem. This reliance on externally supplied initial visualizations could limit the method's generality and applicability to unseen domains where no such visual initialization exists.

\vspace{-0.15cm}
Our approach differs in several key aspects. First, our method autonomously generates conceptual diagrams directly from textual descriptions, without relying on external visual demonstrations or cues, mirroring human ability to construct mental models from language. Second, rather than producing a single visualization, our approach creates multiple chains of intermediate visual states, enabling parallel multi-hypothesis compositional reasoning through evolving diagrams. Third, our diagrams are conceptual, representing relationships and interactions between entities that are visualized with simple shapes rather than realistic depictions. Finally, our method is applicable to any problem expressible in PDDL format (a general language for planning problems) without domain-specific engineering, given only the textual description of the problem. These distinctions collectively enable a more generalizable and flexible form of diagrammatic reasoning, leading to significant performance gains.

\vspace{-0.08cm}
\minisection{Search and Verification Inference Strategies}  
Recent methods improve reasoning in LLMs by structuring inference into explicit intermediate steps~\cite{wei2022chain, kojima2022large} and employing search-based strategies over multiple reasoning paths~\cite{wang2022self, yao2023tree, hao2023reasoning, creswell2022selection, besta2023graph}. For instance, \citet{yao2023tree} propose Tree-of-Thought (ToT), which extends CoT to explore a tree of reasoning paths, and \citet{besta2023graph} introduce Graph-of-Thought (GoT), which structures the reasoning process as a graph, enabling backtracking and aggregation of intermediate reasoning steps. Other works utilize verification and refinement techniques, such as iterative self-reflection and feedback, to enhance reasoning accuracy~\cite{madaan2023self, paul2023refiner, shinn2023reflexion}. Our approach builds upon these methods by integrating a conceptual diagram into each intermediate reasoning step's representation. Moreover, as detailed in Section~\ref{sec:method}, we extend graph-based inference with beam search, significantly reducing the search space and improving performance on long-horizon planning tasks. 

\vspace{-0.08cm}
\minisection{World Modeling with LLMs}  
Recent research explores planning and reasoning using LLMs by implicitly or explicitly constructing world models from textual descriptions~\cite{huang2022language, yao2022react, ahn2022can, dasgupta2022collaborating, driess2023palm, liu2023llm, wang2023voyager}. For instance, \citet{huang2022language} show that LLMs implicitly form textual world models for simple planning tasks, while other works explicitly represent states and transitions for structured reasoning~\cite{yao2022react, dasgupta2022collaborating, liu2023llm}. 
For instance, RAP~\cite{hao2023reasoning}, which we use as a search-based baseline in our evaluation, uses an LLM as both a planning agent and a world modeler, using Monte Carlo Tree Search to simulate future states and rewards that guide its planning. Our work extends these text-based approaches by enabling LMMs to autonomously construct and reason with diagrammatic world models. By defining a visual schema of objects and their statuses using shapes, colors, and spatial arrangements, our method generates diagrams that visually simulate action sequences and state evolutions. This approach provides a more compact representation of relational information, significantly improving performance on complex, multi-step planning tasks, as demonstrated by the substantial accuracy boost on the Blocksworld (hard) domain compared to RAP (78\% vs. 4\%).

%Recent approaches also introduce explicit state-tracking, modeling entity relationships and transitions to enhance reasoning~\cite{liu2023llm, wang2023voyager, driess2023palm}. Our work extends these approaches by enabling LMMs to autonomously construct and reason with diagrammatic world models. Specifically, our method generates conceptual diagrams directly from textual problem descriptions, defining a visual schema of objects and their relationships through abstract shapes, colors, and spatial relationships. Subsequent diagram updates visually simulate action sequences, explicitly tracking object states and interactions. Compared to purely textual world modeling, our diagrammatic approach provides clearer and more accurate representations of spatial and relational information, significantly improving performance on complex, multi-step planning tasks.

\vspace{-0.25cm}
\section{Method}\label{sec:method}
\vspace{-0.2cm}

%Combinatorial planning problems involve finding a valid sequence of actions from an initial state \( s_0 \) to a goal state \( s_g \), given a finite set of possible actions~\cite{garey1979computers}. Such problems are generally NP-hard, and their complexity grows exponentially with solution depth~\cite{bylander1994computational}.

%we translate the resulting action sequence back into PDDL and verify correctness using the VAL plan validator~\cite{howey2004val}.

%We propose Visual Thinking, a training-free and model-agnostic framework, which integrates textual reasoning with model-generated intermediate conceptual diagrams, to enable Large MultiModal Models (LMMs) to solve combinatorial  problems, given text-only problem specifications (initial state, goal, admissible actions). Combinatorial problems ~\cite{combinatorial} involve finding a valid sequence of actions from an initial state \( s_0 \) to a goal state \( s_g \), given a finite set of possible actions. Our framework, built upon the \textit{Graph-of-Thought} approach ~\cite{besta2023graph}, decomposes reasoning into discrete nodes in a structured inference graph, where mulriple chains of multimodal states is simultanously explores toward the goal state.  Below, we detail our method following the stages illustrated in Fig.~\ref{fig:overall}. The full implementation, with prompts and outputs, is available in the supplementary material. An analysis of the prompts and an overview of the code structure are provided in the Appendix.

We propose Visual Thinking, a training-free and model-agnostic framework that integrates textual reasoning with model-generated intermediate conceptual diagrams to enable Large MultiModal Models (LMMs) to solve combinatorial problems, given text-only problem specifications (initial state, goal, admissible actions). Combinatorial problems ~\cite{combinatorial} involve finding a valid sequence of actions from an initial state \( s_0 \) to a goal state \( s_g \), given a finite set of possible actions. Our framework, built upon the Graph-of-Thought (GoT) approach ~\cite{besta2023graph}, decomposes reasoning into discrete nodes in a structured inference graph. Through this graph, multiple chains of multimodal states are simultaneously explored toward the goal state. Below, we detail our method following the stages illustrated in Fig.~\ref{fig:overall}. The full implementation, with full prompts and outputs, is available in the supplementary material. An analysis of the prompts and an overview of the code structure are provided in Appendix Sec.~\ref{sec:appendix_prompts} and Sec.~\ref{sec:appendix_code}, respectively.

\begin{figure}[t] 
    \centering
    \vspace{-0.61cm}
    \includegraphics[width=0.85\columnwidth]{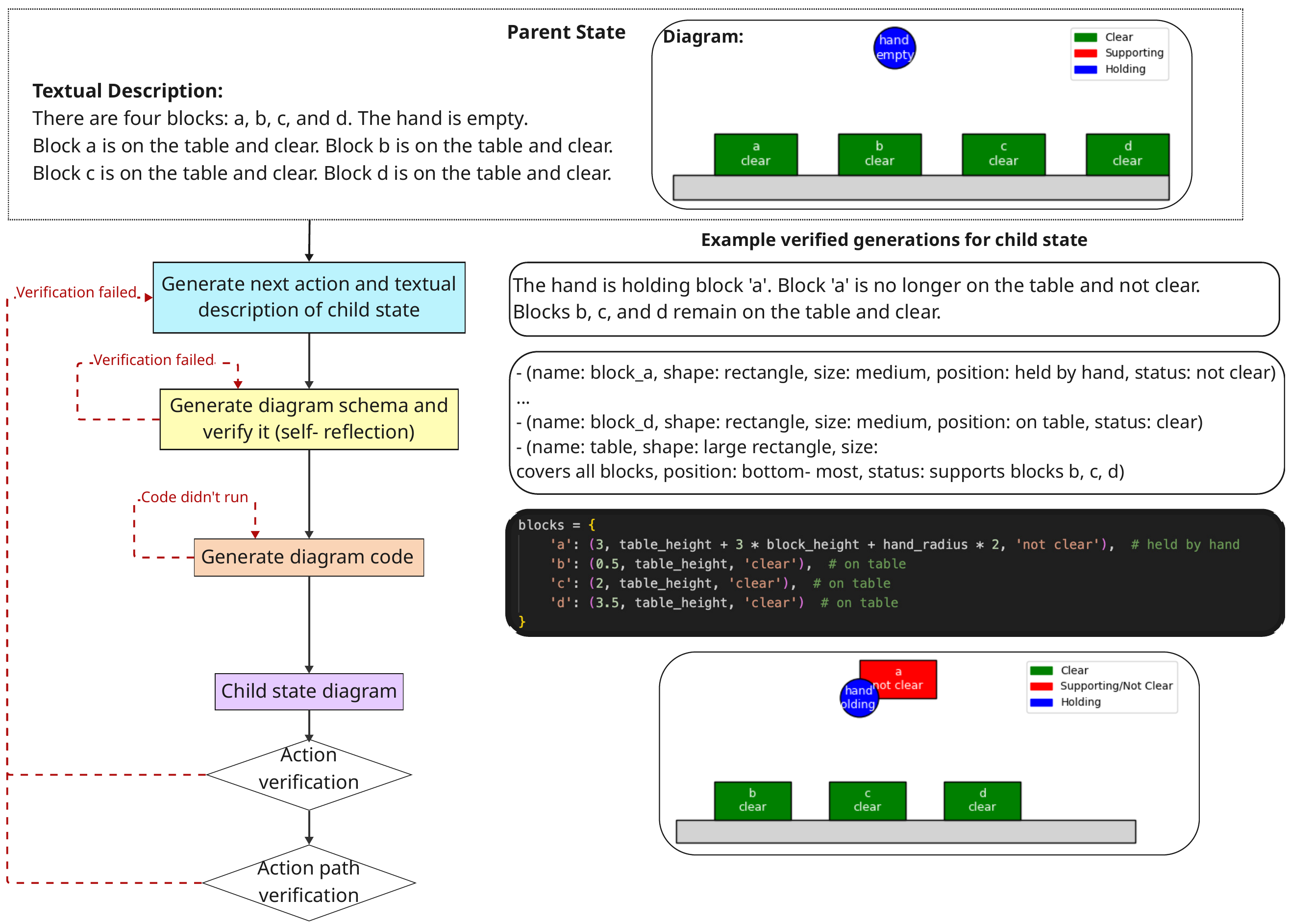}
    \vspace{-0.11cm}
    \caption{\small Child state generation pipeline: the LMM selects an action from the parent node, generates a diagram schema and then an executable diagram code, performs self-reflection and verifies that action chosen does not violate any constraints and action path is feasible (example generations are from Blocksworld domain).}
    \vspace{-0.78cm}
    \label{fig:child_gen}
\end{figure}

\vspace{-0.12cm}
\minisection{1. Generating Initial Conceptual Diagram for Domain} Step 1 in Fig.~\ref{fig:overall}. For each new planning domain, the LMM generates a reference conceptual diagram from a random domain instance. This is done entirely in a zero-shot manner, without manual engineering or external visual cues, conditioned only on the problem's textual description. The process begins with the LMM proposing multiple candidate diagram schemas, which it then verifies by iterating through the objects to confirm the accuracy of their shape, color, status, and relative size and position. The LMM ranks these verified schemas on how clearly they represent object relationships, selecting the top one. Using this schema, the model generates several executable Matplotlib diagram codes. Each rendered diagram is then verified to ensure objects are represented accurately and do not overlap. Finally, the LMM ranks these diagrams based on how effectively they visualize the structure and relationships between objects, and the code for the highest-ranked diagram is cached as a reference for generating the diagram code of the initial and goal states of all instances in the domain. See Fig.~\ref{fig:conceptual_diagram} for example reference diagrams generated for Parking domain.

%For each new planning domain, the LMM generates a reference domain-specific conceptual diagram for a randomly selected instance of the domain. This is done entirely in a zero-shot manner, without manual engineering or external visual cues, conditioned only on the problem's textual description. The LMM first proposes multiple candidate diagram schemas, verifies the,m ny iterating though each object to ensure the object's shpae, color, stattus, and relative size and position is accurate,  and then ranks them clearly they represent object positions and relationships, selecting the top-ranked one. Using this schema and the problem description, we generate multiple executable Matplotlib diagram codes to visualize the domain (Fig. \ref{fig:conceptual_diagram}). Each rendered diagram is verified by the model to objects are accurately represented and nothing overlaps. The LMM then ranks the rendered verified diagrams based on how effectively they visualize the structure and relationships between objects. The highest-ranked diagram is chosen as the `example' domain-specific conceptual diagram, and its code serves as a reference for generating the diagram code for the initial and goal states of all instances in the domain.
\vspace{-0.12cm}
\minisection{2. Initial and Goal State Diagram Generation} Step 2 in Fig.~\ref{fig:overall}.
We begin the inference process of each instance by generating diagrams for the initial state \( s_0 \) and the goal state \( s_g \), conditioned on the domain conceptual diagram code obtained in step 1. When generating the diagram code, the LMM is instructed to adhere to how objects and their statuses are visualized in the reference  diagram, while accurately initializing the objects according to the specific instance.

\vspace{-0.12cm}
\minisection{3. Intermediate Child State Generation Pipeline} Step 3 in Fig.~\ref{fig:overall}.
We denote by \( s_d \) an intermediate state at depth  \( d \) of the graph, represented by a combination of: (i) textual description \( T(s_d) \); (ii) a diagram \( D(s_d) \); and (iii) the action path \( A_{0:d} \) from the initial state \( s_0 \). We iteratively expand the inference graph depth-by-depth in a breadth-first search (BFS) ~\cite{cormen2009introduction} manner, and apply beam search at each depth to select the top‑k candidates for further expansion.

\vspace{-0.1cm}
From each parent state \( s_d \), we sample $n=4$ child states. W.l.o.g., next we describe the generation of a single such child state \( s_{d+1} \) (Fig. \ref{fig:child_gen}). At each node, the LMM first selects the next candidate action \( a_{d+1} \), conditioned on the parent state \( s_d \) (which is represented by its diagram, textual description, and action path from initial state). We then generate the textual description \( T(s_{d+1}) \) of the resulting child state. The candidate action-state pair \((a_{d+1}, T(s_{d+1}))\) is compared to previously generated child states to verify uniqueness. If different, the LMM generates a diagram encoding for \( s_{d+1} \), denoted as  \( E(s_{d+1}) \), which is a structured set of textual statements specifying shapes, sizes, positions, statuses (e.g., colors), and textual identifiers for each object in the state (see Fig.~\ref{fig:child_gen} for an example). The \( E(s_{d+1}) \) undergoes a self-reflection verification to ensure consistency with \( T(s_{d+1}) \) and the action taken, \( a_{d+1} \), if failed we regenerate it. Subsequently, Matplotlib code \( C(s_{d+1}) \) is generated conditioned on \( E(s_{d+1}) \), \( T(s_{d+1}) \), and two example diagram codes: the initial state diagram code \( C(s_0) \) and the parent state diagram code \( C(s_d) \). Code is regenerated if it fails to run.
%
%If either \( E(s_{d+1}) \) or \( C(s_{d+1}) \) fails verification, we internally backtrack within the diagram generation pipeline: specifically, we regenerate the diagram encoding or diagram code, respectively. If repeated attempts still fail, we discard the current child state entirely and prompt the model to select a new action to generate a different child state. 

After generating the child state diagram, we perform two action verifications: (1) a local check that, given the diagram and description of the parent and child state, confirms if the action \( a_{d+1} \) complies with domain constraints; and (2) a global check verifying if the entire action path \( A_{0:d+1} \) is feasible and efficient for reaching the goal state \( s_g \). If either of these checks fails, the child state is marked as invalid.

\vspace{-0.1cm}
\minisection{4. Beam Search} Step 4 in Fig.~\ref{fig:overall}.  
Applying GoT~\cite{besta2023graph} to combinatorial problems by naively expanding all nodes (e.g., BFS) results in exponential growth of the search tree~\cite{russell2010artificial}. To mitigate this, we use a method inspired by beam search ~\cite{lowerre1976harpy}, where first all states at each depth \( d \) are expanded, generating up to \( N \) child states each. The LMM then ranks all candidate child states at depth \( d+1 \) based on their proximity to the goal \( s_g \), selecting only the top \( k=4 \) states for further expansion. This depth-wise pruning mitigates the exponential growth problem. Moreover, this ranking system resembles human problem-solving strategies, where shallow state-specific heuristics are employed to estimate how close intermediate states are to the goal ~\cite{gigerenzer1996reasoning, chronicle2006looks, pizlo2005solving}.

\begin{figure*}[t] 
\vspace{-0.6cm}
    \centering
    \includegraphics[width=1.03\textwidth]{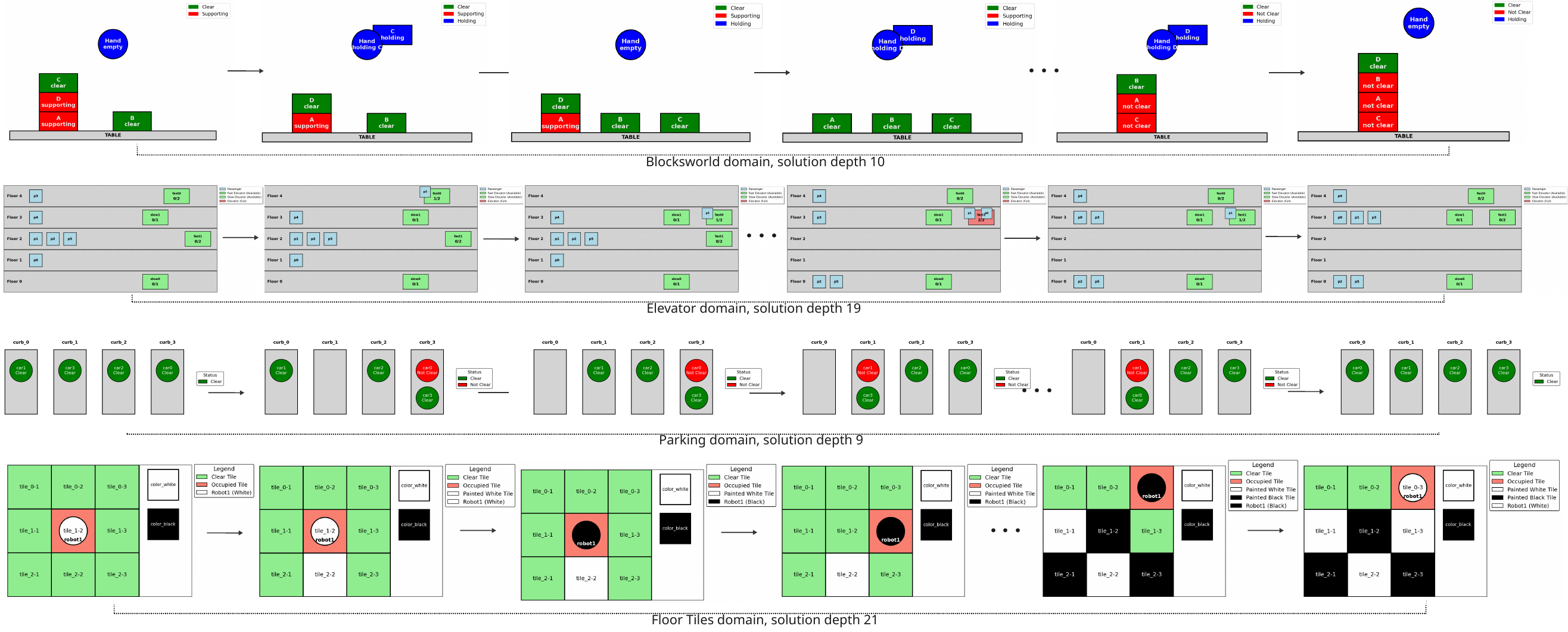}
    \vspace{-0.45cm}
    \caption{\small Sequence of intermediate state diagrams in the correct chain, from the initial to the goal state, for one instance across four evaluated domains: Blocksworld, Elevator, Parking, and Floor Tiles (shown top to bottom).}
    \vspace{-0.55cm}
    \label{fig:state_chains}
\end{figure*}

\vspace{-0.1cm}
\minisection{5. Depth-wise Backtracking} Step 5 in Fig.~\ref{fig:overall}.
Additionally, we implement a depth-wise backtracking mechanism. If all candidate child states at a given depth fail verification, we backtrack to the deepest available ancestor nodes at depth \( d_{\text{max}} \) and attempt new expansions. We allow up to \( B = 2 \) backtracking attempts to any given depth. If all \( B \) attempts at depth \( d_{\text{max}} \) fail, we mark nodes at that depth as invalid and backtrack further to the next deepest available nodes at depth \( d_{\text{max2}} \), where \( d_{\text{max2}} < d_{\text{max}} \).

\vspace{-0.1cm}
\minisection{6. Search Termination} Step 6 in Fig.~\ref{fig:overall}.
The inference process continues iteratively, expanding nodes depth-by-depth, until either the goal state \( s_g \) is reached or a predefined computational budget is exhausted. We set two types of computational limits: (1) a maximum number of generated states (120 states for simpler Blocksworld instances, 450 states for more complex domains), and (2) a maximum depth (28 for simpler Blocksworld instances, 100 for other domains). These limits
ensure computational efficiency in terms of inference time and API usage costs.  If the goal state is not found within these constraints, the search is marked as incomplete. Goal verification occurs at every state expansion, where the LMM compares the diagram and textual description of the current state against those of the goal state.

\vspace{-0.1cm}
In summary, our method leverages structured visual reasoning, self-generated conceptual diagrams, and optimized graph-based inference strategies to efficiently solve combinatorial planning problems with an LMM. Each step of our pipeline is visually illustrated in Figures~\ref{fig:overall},~\ref{fig:conceptual_diagram}, and~\ref{fig:child_gen}.

\vspace{-0.3cm}
\section{Evaluation}\label{sec:eval}
\vspace{-0.25cm}
To evaluate the proposed approach, we conducted experiments on seven different planning domains, including the popular Blocksworld (simple) \cite{gupta1991complexity, valmeekam2023planbench} and Blocksworld (hard) \cite{valmeekam2024planning}, as well as 5 new domains contributed in this work, prepared using a standard repository of PDDL problem instance generators \cite{ipc_generators}. Plan correctness was determined with
VAL~\cite{howey2004val}. Experiments were run on a machine with dual-socket Intel Xeon Gold 5220R processors at 2.2 GHz, 35.75 MB L3, 48 cores per node, 8 nodes total.

\vspace{-0.05cm}

\label{sec:baselines}
\minisection{Baselines} We compare our approach against both single-inference and search-based methods. For single-inference baselines, we evaluated GPT-4o, o1-preview~\cite{openai_o1preview}, and o1-mini~\cite{openai_o1mini}, each prompted  with PDDL instances using templates adapted from~\cite{valmeekam2023planbench}. For search-based methods, we evaluate (i) a baseline variant of Graph-of-Thought (GoT), which performs text-only breadth-first search over LLM-generated states; (ii) Optimized GoT, which adds beam search ($k=4$) to GoT to enable exploring deeper solutions within the compute budget; and (iii) RAP~\cite{hao2023reasoning}, which uses Monte Carlo Tree Search with the LLM generating both the world model and the plan. RAP’s compute budget is specified by iteration count; we adopted RAP\textsuperscript{(20)}, the highest budget reported by the authors in their experiments. RAP was originally evaluated with Llama models; we extended it to GPT-4o for direct comparison. However, RAP’s implementation relies on hard-coded prompts and domain-specific parsers available only for Blocksworld, limiting applicability to other domains. Together, these baselines test the limits of purely text-based search and single inference and provide strong points of comparison for our diagram-based framework.

\begin{table*}[t]
\footnotesize
\centering
\vspace{-0.5cm}
\renewcommand{\arraystretch}{1.05}  % Slightly increase row spacing
\begin{tabular}{lccccccc}
\toprule
\textbf{Analysis} & \footnotesize\makecell{\textbf{Corrects} \\ (Accuracy)}& \footnotesize\makecell{Incorrects} & \footnotesize\makecell{Incompletes} &
\footnotesize\makecell{Avg \\ Depth}  &
\footnotesize\makecell{Max \\ Depth} &
\footnotesize\makecell{Min \\ Depth} &
\footnotesize\makecell{Avg Num\\ States} \\
\midrule
\makecell{Blocksworld (hard)} & 78\% & 10\% & 12\% & 20.15 & 36
 & 18 & 177.7 \\
\specialrule{.4pt}{1pt}{1pt}
Floor Tiles & 36\% & 24\% & 40\% & 15.88 & 25 & 10 & 244 \\
\specialrule{.4pt}{1pt}{1pt}
Parking & 52\%& 40\% & 8\% & 9.6 & 29 & 2 & 149.1 \\
\specialrule{.4pt}{1pt}{1pt}
Tetris & 38\% & 60\% & 2\% & 6.45 & 11 & 4& 46.24 \\
\specialrule{.4pt}{1pt}{1pt}
Elevator & 48\% & 42\% & 10\% & 20.89 & 26 & 16 & 160.30 \\
\specialrule{.4pt}{1pt}{1pt}
Barman & 30\% & 40\% & 30\% & 24.0  & 27 & 22& 210.54 \\
\bottomrule
\end{tabular}
\vspace{-0.18cm}
\caption{\small Analysis of domains. `Corrects': \% instances with correct plans; `Incorrects': \% with incorrect plans; `Incompletes': \% terminated due to state budget; `Avg/Max/Min Depth': avg/max/min number of actions in correct solutions; `Avg Num States': average number of states generated across all instances of the domain.}
\vspace{-0.7cm}
\label{tab:analysis_new}
\end{table*}

\vspace{-0.1cm}
\label{sec:eval:datasets}
\minisection{Evaluated Domains} We evaluated our method on combinatorial planning problems \cite{ghallab2004pddl2} from the International Planning Competition (IPC) \cite{ipc_proceedings}, expressed in PDDL format \cite{mcdermott1998pddl}.  These domains include Blocksworld (simple) and Blocksworld (hard) \cite{gupta1991complexity, valmeekam2024planning}, and five additional IPC domains: Floor Tiles \cite{helmert2000landmarks}, Parking, Tetris \cite{vallati2015automated}, Elevator, and Barman. Instances for new domains were generated using standard publicly available IPC generators~\cite{ipc_generators}. For Blocksworld (which involves stacking and unstacking blocks), we used 500 simple instances from PlanBench~\cite{valmeekam2023planbench} (3–5 blocks) and 50 harder instances (10–20 blocks, following~\cite{valmeekam2024planning}). Floor Tiles features robots painting tiles on a grid; we generated 50 instances with 2–3 rows, 3–5 columns, and 1–2 robots. The Parking domain involves rearranging cars in curbs, with 50 instances using 4-5 curbs and 4-6 cars.  The Tetris domain requires rearranging Tetris tiles on a grid, with 50 instances using $(4\times4)$ or $(6\times6)$ grids. Lastly, the Elevator domain simulates passenger transport in buildings, with 50 instances using 4-5 floors and 10-12 passengers. Figure~\ref{fig:state_chains} shows an example sequence of intermediate state diagrams in the correct plan for a subset of domains. Detailed definitions of each domain are provided in Appendix Sec.~\ref{sec:appendix_domains}.

%For each domain, we generated instances using publicly available instance generators \cite{ipc_generators}. The Blocksworld domain involves stacking and unstacking blocks. We used the 500 simple instances in PlanBench \cite{valmeekam2023planbench} with 3-5 blocks and 50 additional hard instances with 10-20 blocks (same instance complexity as in \cite{valmeekam2024planning}). The Floor Tiles domain features robots painting tiles on a grid. We generated 50 PDDL instances for this domain with varying grid sizes (2-3 rows, 3-5 columns) and 1-2 robots. The Parking domain involves rearranging cars in curbs, with 50 instances using 4-5 curbs and 4-6 cars. The Tetris domain requires rearranging Tetris tiles on a grid, with 50 instances using $(4\times4)$ or $(6\times6)$ grids. Lastly, the Elevator domain simulates passenger transport in buildings, with 50 instances using 4-5 floors and 10-12 passengers. Refer to Fig. \ref{fig:state_chains} for an exmaple sequence of state diagrams in the correct chain of a subset of evaluated domains. Detailed definitions of each domain are provided in Appendix Sec.~\ref{}. 

\vspace{-0.1cm}
\minisection{Translating PDDL to Natural Language and Back} Our method, Visualizing Thought, operates on the natural language description of combinatorial problems. To enable this, we first translate each PDDL domain—the rules and allowed actions—into natural language using a manually engineered five-shot prompt covering five different domains. Each instance, which specifies the initial and goal states, is translated with a one-shot prompt. Our proposed approach then runs entirely on this text representation. After solving the problem, the model’s natural language action sequence is translated back into PDDL using a one-shot prompt containing a random (incorrect) plan with correct PDDL syntax. The resulting PDDL plan is then evaluated for correctness using VAL.

\vspace{-0.2cm}
\subsection{Results and Analysis}
\vspace{-0.1cm}
\label{sec:eval:results}

Our main results are presented in Table \ref{tab:all_results}, comparing Visual Thinking (with GPT-4o) against leading reasoning models (o1-preview, o1-mini) and strong search-based methods (GoT, Optimized GoT, RAP). Visual Thinking substantially improves over base GPT-4o and consistently outperforms all search-based approaches across domains. On Blocksworld (simple), our method achieves 90.2\% accuracy, surpassing GoT, Optimized GoT, and RAP, though slightly trailing o1-preview (97.8\%). We conjecture this gap is due to the smaller number of entities, which make the world state easier to track and update in text, and the shallow solution depths, which make these instances easier to solve in a single pass. In contrast, on harder domains, including Blocksworld (hard) and the five new domains, our approach shows considerable, generalizable gains. For example, on Blocksworld (hard) we achieve 78\%, compared to 23.65\% for o1-preview and 4\% for RAP. This trend holds across other domains. Standard GoT often fails completely (e.g., 0\% on Floor Tiles and Tetris) due to combinatorial explosion exhausting the budget, and while Optimized GoT mitigates this with beam search, its performance still lags well behind our visual approach. These findings highlight how diagram-based reasoning enables models to capture and analyze complex relational structures more efficiently than purely textual inference.

\vspace{-0.1cm}

Table~\ref{tab:analysis_new} provides further insight into our method's performance on the more challenging domains. Despite significantly deeper solution paths (instances were designed with solution paths of up to 40), our method successfully generates correct plans with as many as 36 sequential actions (Blocksworld (hard)). The primary limitation of our method observed in these experiments is the number of incomplete searches (e.g., 40\% incomplete in Floor Tiles, 30\% in Barman), which arise either when invalid actions are later rejected by local verification using state diagrams, or when inefficient actions that fail to advance toward the goal are pruned by the global check, both leading to exhaustion of the computational budget. We also observe the highest incorrect rate (60\%) on the Tetris domain, primarily due to inherently high branching factor in this domain (up to 24 possible actions per state) and complex action parameterization—each action can require up to 7 parameters detailing positions of all sub-tiles, compared to simpler domains like Blocksworld, where actions typically require only 1-2 parameters. Conversely, our largest margin over o1-preview occurs in the Barman domain (30\% vs. 10\%), likely because diagrammatic representations capture the high number of object statuses and interactions per state  in this domain more effectively than text alone.

\vspace{-0.1cm}

\minisection{Model Generalization} To assess the generalizability of our framework, we evaluated it on other state-of-the-art LMMs using 50 instances from the Blocksworld (simple) domain. With Llama 4, our method increased accuracy from a baseline of 10\% (single inference) to 74\%, an over 7x improvement in accuracy. The improvement was also observed using Claude 3.5 Sonnet, where accuracy increased from 54.8\% (using zero shot single inference as reported in~\cite{valmeekam2024planning}) to 98\%, achieving state-of-the-art performance on this benchmark. These substantial gains demonstrate that the benefits of our approach are not tied to a specific model architecture but stem from the fundamental advantage of using model-generated conceptual diagrams as a reasoning medium.

\vspace{-0.1cm}

\minisection{Runtime and Cost Analysis} We analyzed the runtime and API costs of our method and all other search-based baselines on 20 instances per domain. On the Blocksworld (simple) domain, our method had a median runtime of 381 ($\sim$6 minutes) seconds and a median cost of \$1.04 per instance. For more complex domains, the median runtime was 1038 seconds ($\sim$17 minutes) with median cost of \$2.98 per instance. Our approach is significantly more efficient than other search-based methods. On average, it was 31\% faster and 36\% cheaper than the GoT baseline, and 46\% faster and 52\% cheaper than RAP\textsuperscript{20}+GPT-4o across all domains, while achieving substantially higher accuracy. Compared to the text-only Optimized GoT, incorporating diagrams added 213 seconds in latency and \$0.71 on average, measured across all domains, but this overhead yielded a 30 percentage point accuracy gain. 

\vspace{-1cm}
\begin{table*}[t]
\centering
\footnotesize
\renewcommand{\arraystretch}{1.}  % Slightly increase row spacing
\vspace{-0.5cm}
\begin{tabular}{lcccccc}
\toprule
\textbf{Ablation} & \footnotesize\makecell{\textbf{Corrects} \\ (Accuracy)}& \footnotesize\makecell{Incorrects} & \footnotesize\makecell{Incompletes} &
\footnotesize\makecell{Avg \\ Depth}  &
\footnotesize\makecell{Max\\ Depth} &
\footnotesize\makecell{Avg Num \\States} \\
\midrule
GPT-4o + Visual Thinking  & 90.2\% & 6.4\% & 3.4\% & 10.07 & 28 & 38.89 \\
\specialrule{1pt}{1pt}{1pt}
No Diagram \footnotesize{(Optimized GoT)} & 58\% & 36\% & 6\% &  8.28& 18 & 25.86 \\
\specialrule{.4pt}{1pt}{1pt}
No Diagram Schema & 72\%& 20\% & 8\% & 7.41 & 18  & 46.2 \\
\specialrule{.4pt}{1pt}{1pt}
No Code Execution & 24\% & 66\% & 10\% & 7.82 & 22  & 32.7 \\
\specialrule{1pt}{1pt}{1pt}
1-Branching Factor  & 52\% & 4\% & 44\% & 6.38 & 16  & 23.9  \\
\specialrule{.4pt}{1pt}{1pt}
2-Branching Factor & 70\% & 6\% & 24\% & 8.29 & 24  &  26.27 \\
\specialrule{1pt}{1pt}{1pt}
No Backtracking & 62\% & 4\%& 34\% & 6.06 & 14  & 22.12 \\
\specialrule{.4pt}{1pt}{1pt}
No Beam Search  & 72\%& 4\% & 24\% & 6.61& 12 & 58.31 \\
\bottomrule
\end{tabular}
\vspace{-0.18cm}
\caption{\small Ablation Study Results. `Corrects' is our main accuracy metric. See Tab. \ref{tab:analysis_new} for columns notation.}
\vspace{-0.7cm}
\label{tab:ablation}

\end{table*}

\vspace{0.75cm}
\subsection{Ablation Studies}
\vspace{-0.1cm}
\label{sec:eval:abl}

To systematically evaluate the contributions of different components of our framework, we conducted ablation studies on 50 instances from the Blocksworld (simple) domain. We examined the impact of state diagrams, diagram schema, diagram code execution, different branching factors, and inference optimizations (beam search and backtracking). Table~\ref{tab:ablation} summarizes the results of these experiments.

\vspace{-0.1cm}

\minisection{Impact of Various Components of State Diagram Generation} We first evaluated the role of state diagrams by removing them entirely from the inference pipeline (“No Diagram”), yielding a text-only optimized Graph-of-Thought approach. This caused accuracy to drop from 90.2\% to 58\%, underscoring the critical role diagrams play in succinctly representing relational information. Moreover, the average solution depth of correctly solved instances decreased from 10.07 to 8.28, indicating that without diagrams, the model struggled on more complex problems requiring deeper reasoning. In a second experiment, we removed the diagram schema (“No Diagram Schema”) from the child-state generation pipeline, instead inferring diagram code directly from textual descriptions of states. Accuracy dropped from 90.2\% to 72\%, a smaller decline than removing diagrams entirely—showing that diagram schemas further help extract relational information from text, enabling more accurate diagram generation.

%The average number of states generated increased notably (46.2 vs. 38.89).
% , while the percentage of validated states decreased (44.14\% vs. 56.45\%), indicating that without diagram schema generation step, the model frequently generated invalid diagrams, which were subsequently rejected during verification. 

\vspace{-0.13cm}

Finally, we tested removing the rendered diagrams, providing only the Matplotlib code ("No Code Execution") when generating the next action. This resulted in the most significant performance drop (90.2\% to 24\%), even below the GPT-4o baseline (35.5\%), clearly demonstrating that even though the code encodes the same spatial and relational information as the diagram, the way this information and the interdependencies between objects are represented is crucial for model performance. Using the diagram code directly distracts the model and impairs reasoning, aligning with prior findings that extraneous details negatively impact model performance~\cite{shi2023large,liu2023lost}. These results reinforce the importance of diagrams as compact, intuitive representations that facilitate rapid perceptual inference and clear relational reasoning~\cite{larkin1987diagrammatic,tversky2011visualizing,hegarty2004mechanical}, and that it is the representation of information in a multi-dimensional (2D, color-encoded) format that significantly aids understanding the interdependencies and reasoning in models.

\vspace{-0.13cm}

\minisection{Impact of Branching Factor}  We next investigated the effect of branching factor of the inference graph (the number of candidate child states generated per state) on performance. Reducing the branching factor from 4 (our default) to 2 ("2-Branching Factor") decreased performance from 90.2\% to 70\%, primarily due to a sharp increase in incomplete searches (24\% vs. 3.4\%). This suggests that exploring the third or fourth candidate child states, generated at higher temperatures, is frequently necessary to find the correct solution path. Thus, reducing branching factor limits diversity in candidate state generations, leading to more incomplete searches. Additionally, the average depth of correctly solved instances decreased from 10.07 to 8.29, indicating difficulty solving problems with deeper solution depths. Further reducing the branching factor to 1 ("1-Branching Factor"), effectively converting our graph-based inference into a multimodal Chain-of-Thought approach (with diagrams), caused performance to drop even further to 52\%. Despite this decline, performance remained well above the GPT-4 baseline (35.5\%), underscoring the value of diagrams in improving LMM reasoning even without extensive search.

\vspace{-0.13cm}

\minisection{Impact of Inference Optimizations} Finally, we evaluated the importance of our inference optimizations (backtracking and beam search) on top of the multimodal inference graph. Removing backtracking (“No Backtracking”)—yielding a tree-of-thought method with beam search—reduced accuracy from 90.2\% to 62\%, primarily due to a sharp rise in incomplete searches (34\% vs. 3.4\%). This occurs because LMM verification steps occasionally produce false negatives, incorrectly invalidating correct states. Without backtracking, the model cannot recover from these errors, leading to incomplete searches as no validated nodes remain at the frontier of the search graph for further expansion. 

\vspace{-0.13cm}

Similarly, removing beam search (“No Beam Search”) lowered accuracy to 72\%, with the incomplete search rate increasing to 24\%. In this case, incompletes stem from exponential growth in the search space, causing the model to exhaust its computational budget (i.e., the maximum number of states generated) before reaching the goal. Indeed, the average number of generated states increased significantly (58.31 vs. 38.89), underscoring the critical role of beam search in managing combinatorial explosion. Both optimizations are essential for solving deeper combinatorial problems, as shown by the reduced average correct solution depth without them (6.61 without beam search, 6.06 without backtracking, vs. 10.07 with both). These results demonstrate that backtracking and beam search are complementary and crucial for efficient graph-based combinatorial planning.

\vspace{-0.25cm}
\section{Conclusion}
\vspace{-0.15cm}
\label{sec:conclusion}
\minisection{Contributions} In this paper, we introduced Visual Thinking, a framework that enables LMMs to solve combinatorial problems by reasoning with conceptual diagrams alongside text. Our contributions are: (i) a cognitively inspired method that autonomously generates conceptual diagrams directly from natural language problem descriptions, requiring no human input for new domains; (ii) a multimodal Graph-of-Thought framework that structures reasoning as sequences of intermediate textual and visual states, integrating beam search and backtracking for efficient long-horizon search; (iii) extensive empirical evidence showing substantial performance gains over single-inference LMMs, specialized reasoning models, and strong search-based baselines across various planning benchmarks; and (iv) ablation results demonstrating that representation of information is critical—reasoning improves when relational information and interdependencies are encoded in diagrams, not merely present in text or code format.

\vspace{-0.13cm}

\label{sec:limitations}

\minisection{Limitations} As with any search-based inference method, our framework incurs additional computational cost and inference time to explore multiple reasoning trajectories and also to generate visual representations. However, these overheads are manageable in practice; importantly, our approach remains more efficient than prior text-only search methods such as Graph-of-Thought and RAP. Another limitation concerns scope: Visual Thinking was applied to combinatorial planning problems—a core area of computer science with significant real-world applications, such as warehouse optimization, logistics, and scheduling—chosen because their state-based structure makes diagram progression more straightforward. Future work involves extending conceptual diagram generation to more open-ended problems.

\vspace{-0.13cm}
\label{sec:future_work}
%\minisection{Future Work} This work demonstrates a path for LMMs to move beyond purely textual reasoning toward a more powerful, human-like process that integrates visual abstractions. Future work can extend this framework from combinatorial planning to more abstract domains, enabling LMMs to generate multimodal outputs such as software architecture diagrams, figures to visualize scientific hypotheses and causal dependencies, or tailored visual aids for educational contexts. Such conceptual diagrams can not only enhance models' performance but also enhance human AI interaction (e.g., easier verification of code performance using generated architecture diagrams). This ability to reason and communicate multimodally is crucial for building AI that can tackle complex scientific, creative, and planning tasks.

\minisection{Future Work} This work demonstrates a path for LMMs to move beyond purely textual reasoning toward a more powerful, human-like process that integrates visual abstractions. Future work can extend this framework beyond combinatorial planning to more abstract domains, enabling LMMs to produce multimodal outputs such as software architecture diagrams, figures visualizing scientific hypotheses and causal dependencies, or tailored visual aids for educational contexts. Such conceptual diagrams can enhance both model performance and human–AI interaction (e.g., easier verification of code behavior through generated architecture diagrams). This ability to reason and communicate about complex structures multimodally is essential for building AI capable of tackling scientific, creative, and planning tasks.

%\section{Reproducibility Statement}
%\vspace{-0.2cm}
%An anonymous supplementary repository includes the full inference code, prompt templates, and scripts used in our experiments, along with the PDDL instances of the new planning benchmark created for the five new domains and sample run results. The end-to-end pipeline is specified in Section~\ref{sec:method} and summarized in Figure~\ref{fig:overall}. Dataset sources and instance-generation parameters are detailed in Section~\ref{sec:eval:datasets}; formal task and action definitions required to regenerate the PDDL domains are provided in Appendix~\ref{sec:appendix_domains}; and prompt templates are listed in Appendix~\ref{sec:appendix_prompts}. Experimental setup, baselines, and evaluation protocols are described in Sections~\ref{sec:eval} and \ref{sec:eval:results}, with ablations in Section~\ref{sec:eval:abl} and comprehensive results in Tables~\ref{tab:all_results} and~\ref{tab:ablation}. Code organization and configuration files are documented in Appendix~\ref{sec:appendix_code}.

\vspace{-0.35cm}
\medskip
\bibliography{iclr2026_conference}

\begin{thebibliography}{65}
\providecommand{\natexlab}[1]{#1}
\providecommand{\url}[1]{\texttt{#1}}
\expandafter\ifx\csname urlstyle\endcsname\relax
  \providecommand{\doi}[1]{doi: #1}\else
  \providecommand{\doi}{doi: \begingroup \urlstyle{rm}\Url}\fi

\bibitem[ipc()]{ipc_generators}
Pddl instance generators.
\newblock URL \url{http://ipc.icaps-conference.org/domains/}.

\bibitem[ipc(1998--Present)]{ipc_proceedings}
\emph{Proceedings of the International Planning Competition}, 1998--Present.
\newblock URL \url{http://ipc.icaps-conference.org/}.
\newblock This citation refers to the proceedings and general information about the International Planning Competition (IPC).

\bibitem[Ahn et~al.(2022)Ahn, Brohan, Brown, Chebotar, Cortes, David, Du, Duong, Edunov, Gomez, et~al.]{ahn2022can}
Michael Ahn, Noah Brohan, Noah Brown, Yevgenii Chebotar, Omar Cortes, Byron David, Chelsea Du, Keerthana Duong, Sergey Edunov, A{\'a}ron Gomez, et~al.
\newblock Can language models learn from explanations in context?
\newblock \emph{arXiv preprint arXiv:2204.02329}, 2022.

\bibitem[Anthropic(2024{\natexlab{a}})]{claude3}
Anthropic.
\newblock Introducing the next generation of claude, 2024{\natexlab{a}}.
\newblock URL \url{https://www.anthropic.com/news/claude-3-family}.

\bibitem[Anthropic(2024{\natexlab{b}})]{sonnet}
Anthropic.
\newblock Claude 3.5 sonnet, 2024{\natexlab{b}}.
\newblock URL \url{https://www.anthropic.com/news/claude-3-5-sonnet}.

\bibitem[Battaglia et~al.(2013)Battaglia, Hamrick, and Tenenbaum]{battaglia2013simulation}
Peter~W Battaglia, Jessica~B Hamrick, and Joshua~B Tenenbaum.
\newblock Simulation as an engine of physical scene understanding.
\newblock \emph{Proceedings of the national academy of sciences}, 110\penalty0 (45):\penalty0 18327--18332, 2013.

\bibitem[Besta et~al.(2023)Besta, Gerstenberger, Rausch, Fischer, Lehmann, Podstawski, Huschenbett, B{\"o}ttger, and Kersting]{besta2023graph}
Maciej Besta, Nils Gerstenberger, Robert Rausch, Tim Fischer, Maximilian Lehmann, Kamil Podstawski, Christoph Huschenbett, Andreas B{\"o}ttger, and Kristian Kersting.
\newblock Graph of thoughts: Solving elaborate problems with large language models.
\newblock 2023.

\bibitem[Borazjanizadeh \& Piantadosi(2024)Borazjanizadeh and Piantadosi]{borazjanizadeh2024reliable}
Nasim Borazjanizadeh and Steven~T Piantadosi.
\newblock Reliable reasoning beyond natural language.
\newblock \emph{arXiv preprint arXiv:2407.11373}, 2024.

\bibitem[Borazjanizadeh et~al.(2024)Borazjanizadeh, Herzig, Darrell, Feris, and Karlinsky]{borazjanizadeh2024navigating}
Nasim Borazjanizadeh, Roei Herzig, Trevor Darrell, Rogerio Feris, and Leonid Karlinsky.
\newblock Navigating the labyrinth: Evaluating and enhancing llms' ability to reason about search problems.
\newblock \emph{arXiv preprint arXiv:2406.12172}, 2024.

\bibitem[Chronicle et~al.(2006)Chronicle, MacGregor, Ormerod, and Burr]{chronicle2006looks}
Edward~P Chronicle, James~N MacGregor, Thomas~C Ormerod, and Alistair~H Burr.
\newblock It looks easy! heuristics for combinatorial optimization problems.
\newblock \emph{The Quarterly Journal of Experimental Psychology}, 59\penalty0 (4):\penalty0 783--800, 2006.

\bibitem[Clark(1996)]{clark1996using}
Herbert~H Clark.
\newblock \emph{Using language}.
\newblock Cambridge university press, 1996.

\bibitem[Clottes(2008)]{clottes2008cave}
Jean Clottes.
\newblock \emph{Cave art}.
\newblock Phaidon Press, 2008.

\bibitem[Cobbe et~al.(2021)Cobbe, Kosaraju, Bavarian, Chen, Jun, Kaiser, Plappert, Tworek, Hilton, Nakano, Hesse, and Schulman]{cobbe2021gsm8k}
Karl Cobbe, Vineet Kosaraju, Mohammad Bavarian, Mark Chen, Heewoo Jun, Lukasz Kaiser, Matthias Plappert, Jerry Tworek, Jacob Hilton, Reiichiro Nakano, Christopher Hesse, and John Schulman.
\newblock Training verifiers to solve math word problems.
\newblock \emph{arXiv preprint arXiv:2110.14168}, 2021.

\bibitem[Cormen et~al.(2009)Cormen, Leiserson, Rivest, and Stein]{cormen2009introduction}
Thomas~H Cormen, Charles~E Leiserson, Ronald~L Rivest, and Clifford Stein.
\newblock \emph{Introduction to algorithms}.
\newblock MIT press, 3rd edition, 2009.
\newblock Breadth-first search algorithm described in Chapter 22.2.

\bibitem[Creswell et~al.(2023)Creswell, Weber, Upadhyay, Lampson, Uesato, Kohli, and Das]{creswell2023selection}
Antoni Creswell, Lisa Weber, Yogesh Upadhyay, Nicholas Lampson, Jonathan Uesato, Pushmeet Kohli, and Rishabh Das.
\newblock Selection-inference: Exploiting large language models for interpretable logical reasoning.
\newblock In \emph{Thirty-seventh Conference on Neural Information Processing Systems}, 2023.

\bibitem[Creswell et~al.(2022)Creswell, Shanahan, and Higgins]{creswell2022selection}
Antonia Creswell, Murray Shanahan, and Irina Higgins.
\newblock Selection-inference: Exploiting large language models for interpretable logical reasoning.
\newblock \emph{arXiv preprint arXiv:2205.09712}, 2022.

\bibitem[Dasgupta et~al.(2022)Dasgupta, Murugesan, Jiang, Murthi, Zhu, Yang, Wong, Banerjee, Raffin, Zeng, et~al.]{dasgupta2022collaborating}
Ishita Dasgupta, Karmanya Murugesan, Orion Jiang, Karthik Murthi, Michael Zhu, Cunjun Yang, Derek Wong, Sriram Banerjee, Antonin Raffin, Andy Zeng, et~al.
\newblock Collaborating with language models for embodied reasoning.
\newblock \emph{arXiv preprint arXiv:2207.05608}, 2022.

\bibitem[Driess et~al.(2023)Driess, Chowdhery, Schrittwieser, Laskin, Shah, Castro, Clark, Cortés, Aherne, et~al.]{driess2023palm}
Danny Driess, Aakanksha Chowdhery, Josef Schrittwieser, Ferenc Laskin, Michael~and, Ayzaan Shah, Carolina Castro, Nadine Clark, Omar Cortés, Fergal Aherne, et~al.
\newblock Palm-e: An embodied multimodal language model.
\newblock \emph{arXiv preprint arXiv:2303.03378}, 2023.

\bibitem[Dziri et~al.(2023)Dziri, Chaudhuri, Manning, Liang, and Hashimoto]{dziri2023faith}
Nouha Dziri, Siddhartha Chaudhuri, Christopher~D Manning, Percy Liang, and Tatsunori Hashimoto.
\newblock Faith and fate: Limits of transformers on compositionality.
\newblock In \emph{International Conference on Machine Learning}, pp.\  10454--10481. PMLR, 2023.

\bibitem[Gentner \& Stevens(2001)Gentner and Stevens]{gentner2001mental}
Dedre Gentner and Albert~L Stevens.
\newblock Mental models, psychology of.
\newblock \emph{International encyclopedia of the social \& behavioral sciences}, pp.\  9609--9613, 2001.

\bibitem[Ghallab et~al.(2004)Ghallab, Fox, Long, Smith, Cremers, and Hoffmann]{ghallab2004pddl2}
Malik Ghallab, Maria Fox, Derek Long, David Smith, Anthony Cremers, and J{\"o}rg Hoffmann.
\newblock \emph{PDDL2. 2-the language for describing planners}.
\newblock Yale University Department of Computer Science New Haven, CT, USA, 2004.

\bibitem[Gigerenzer \& Goldstein(1996)Gigerenzer and Goldstein]{gigerenzer1996reasoning}
Gerd Gigerenzer and Daniel~G Goldstein.
\newblock Reasoning the fast and frugal way: Models of bounded rationality.
\newblock \emph{Psychological review}, 103\penalty0 (4):\penalty0 650, 1996.

\bibitem[Gupta \& Nau(1991)Gupta and Nau]{gupta1991complexity}
Neelima Gupta and Dana~S Nau.
\newblock The complexity of blocks-world planning.
\newblock \emph{AAAI}, pp.\  640--646, 1991.
\newblock Relevant sentence: "The complexity of blocks-world planning is examined in detail.".

\bibitem[Hao et~al.(2023)Hao, Gu, Ma, Hong, Wang, Wang, and Hu]{hao2023reasoning}
Shibo Hao, Yi~Gu, Haodi Ma, Joshua~Jiahua Hong, Zhen Wang, Daisy~Zhe Wang, and Zhiting Hu.
\newblock Reasoning with language model is planning with world model.
\newblock \emph{arXiv preprint arXiv:2305.14992}, 2023.

\bibitem[Hegarty(2004)]{hegarty2004mechanical}
Mary Hegarty.
\newblock Mechanical reasoning by mental simulation.
\newblock \emph{Trends in cognitive sciences}, 8\penalty0 (6):\penalty0 280--285, 2004.

\bibitem[Helmert(2000)]{helmert2000landmarks}
Malte Helmert.
\newblock Landmarks in planning.
\newblock In \emph{Principles of knowledge representation and reasoning}, pp.\  97--106, 2000.
\newblock Relevant sentence: "Landmarks are propositions that must be true in every plan that solves a given planning problem.".

\bibitem[Hendrycks et~al.(2021)Hendrycks, Burns, Kadavath, Arora, Basart, Tang, Song, and Steinhardt]{hendrycks2021measuring}
Dan Hendrycks, Collin Burns, Saurav Kadavath, Akul Arora, Steven Basart, Eric Tang, Dawn Song, and Jacob Steinhardt.
\newblock Measuring mathematical problem solving with the math dataset.
\newblock \emph{arXiv preprint arXiv:2103.03874}, 2021.

\bibitem[Howey et~al.(2004)Howey, Long, and Fox]{howey2004val}
Richard Howey, Derek Long, and Maria Fox.
\newblock Val: automatic plan validation, continuous effects and mixed initiative planning using pddl.
\newblock In \emph{Proceedings 16th IEEE International Conference on Tools with Artificial Intelligence}, pp.\  294--301. IEEE, 2004.

\bibitem[Hu et~al.(2024)Hu, Shi, Fu, Roth, Ostendorf, Zettlemoyer, Smith, and Krishna]{hu2024visual}
Yushi Hu, Weijia Shi, Xingyu Fu, Dan Roth, Mari Ostendorf, Luke Zettlemoyer, Noah~A Smith, and Ranjay Krishna.
\newblock Visual sketchpad: Sketching as a visual chain of thought for multimodal language models.
\newblock \emph{arXiv preprint arXiv:2406.09403}, 2024.

\bibitem[Huang et~al.(2022)Huang, Xia, Shah, Zeng, Janner, Levine, Finn, Garg, Dragan, and Song]{huang2022language}
Wenlong Huang, Fei Xia, Ayzaan Shah, Andy Zeng, Michael Janner, Sergey Levine, Chelsea Finn, Animesh Garg, Anca~D Dragan, and Liwei Song.
\newblock Language models as zero-shot planners: Extracting actionable knowledge for embodied agents.
\newblock In \emph{International Conference on Machine Learning}, pp.\  8844--8867. PMLR, 2022.

\bibitem[Johnson-Laird(1983)]{johnson1983mental}
Philip~Nicholas Johnson-Laird.
\newblock \emph{Mental models}.
\newblock Harvard University Press, 1983.

\bibitem[Kojima et~al.(2022)Kojima, Sano, Yan, Furukawa, Sadaie, and Yanai]{kojima2022large}
Takeshi Kojima, Shixiang Sano, Hideaki Yan, Keizo Furukawa, Jun Sadaie, and Kunihiko Yanai.
\newblock Large language models are zero-shot reasoners.
\newblock In \emph{Advances in neural information processing systems}, 2022.

\bibitem[Larkin \& Simon(1987)Larkin and Simon]{larkin1987diagrammatic}
Jill~H Larkin and Herbert~A Simon.
\newblock Why a diagram is (sometimes) worth ten thousand words.
\newblock \emph{Cognitive science}, 11\penalty0 (1):\penalty0 65--99, 1987.

\bibitem[Liu et~al.(2023{\natexlab{a}})Liu, Zhang, Chen, Gao, Zhang, Zhou, Jiang, Wang, and Chen]{liu2023llm}
Keming Liu, Tianhua Zhang, Yujie Chen, Jia Gao, Wei Zhang, Xiang Zhou, Tao Jiang, Junsheng Wang, and Tie-Yan Chen.
\newblock {LLM+}: Augmenting language models with explicit state representations.
\newblock \emph{arXiv preprint arXiv:2305.06975}, 2023{\natexlab{a}}.

\bibitem[Liu et~al.(2023{\natexlab{b}})Liu, Shen, Zhang, and Tang]{liu2023lost}
Nelson~F Liu, Kevin Shen, Keyulu Zhang, and Jian Tang.
\newblock Lost in the middle: How language models use long contexts.
\newblock \emph{arXiv preprint arXiv:2307.03172}, 2023{\natexlab{b}}.

\bibitem[Lowerre(1976)]{lowerre1976harpy}
Bruce~T Lowerre.
\newblock The harpy speech recognition system.
\newblock \emph{Ph.D. Dissertation, Carnegie Mellon University}, 1976.
\newblock Often cited as the origin of Beam Search, though the term "beam search" wasn't explicitly used in this dissertation. Lowerre describes a similar approach for reducing search space in speech recognition.

\bibitem[Madaan et~al.(2023)Madaan, Khapra, Tsvetkov, and Salakhutdinov]{madaan2023self}
Aman Madaan, Prateek Khapra, Yulia Tsvetkov, and Ruslan Salakhutdinov.
\newblock Self-refine: Iterative refinement with self-feedback.
\newblock \emph{arXiv preprint arXiv:2303.17651}, 2023.

\bibitem[Mayer(2002)]{mayer2002multimedia}
Richard~E Mayer.
\newblock \emph{Multimedia learning}.
\newblock Cambridge university press, 2002.

\bibitem[McDermott et~al.(1998)McDermott, Ghallab, Howe, Knoblock, Ram, Veloso, Weld, and Wilkins]{mcdermott1998pddl}
Drew McDermott, Malik Ghallab, Adele Howe, Craig Knoblock, Ashwin Ram, Manuela Veloso, Daniel Weld, and David Wilkins.
\newblock Pddl--the planning domain definition language.
\newblock Technical Report TR-98-003/DCS TR-1165, Yale Center for Computational Vision and Control, 1998.

\bibitem[Meta(2025)]{Llama4}
Meta.
\newblock Llama 4, 2025.
\newblock URL \url{https://ai.meta.com/blog/llama-4-multimodal-intelligence/}.

\bibitem[OpenAI(2023)]{OpenAI2023GPT4TR}
OpenAI.
\newblock Gpt-4 technical report.
\newblock \emph{ArXiv}, abs/2303.08774, 2023.

\bibitem[OpenAI(2024{\natexlab{a}})]{openai_gpt4o}
OpenAI.
\newblock Hello gpt-4o, 2024{\natexlab{a}}.
\newblock URL \url{https://openai.com/index/hello-gpt-4o/}.

\bibitem[OpenAI(2024{\natexlab{b}})]{openai_o1mini}
OpenAI.
\newblock Openai o1-mini, 2024{\natexlab{b}}.
\newblock URL \url{https://openai.com/index/openai-o1-mini-advancing-cost-efficient-reasoning/}.

\bibitem[OpenAI(2024{\natexlab{c}})]{openai_o1preview}
OpenAI.
\newblock Introducing openai o1-preview, 2024{\natexlab{c}}.
\newblock URL \url{https://openai.com/index/introducing-openai-o1-preview/}.

\bibitem[Paul et~al.(2023)Paul, Laskin, Chiang, Jiang, Du, Levine, Grosse, Yao, et~al.]{paul2023refiner}
Sreejan~Kumar Paul, Michael Laskin, Kevin Chiang, Yi~Jiang, Nan Du, Sergey Levine, Roger Grosse, Yuan Yao, et~al.
\newblock Refiner: Reasoning feedback on intermediate representations.
\newblock \emph{arXiv preprint arXiv:2304.01904}, 2023.

\bibitem[Pinker \& Bloom(1990)Pinker and Bloom]{pinker1990natural}
Steven Pinker and Paul Bloom.
\newblock Natural language and natural selection.
\newblock \emph{Behavioral and Brain Sciences}, 13\penalty0 (4):\penalty0 707--727, 1990.

\bibitem[Pizlo \& Li(2005)Pizlo and Li]{pizlo2005solving}
Zygmunt Pizlo and Zheng Li.
\newblock Solving 756 combinatorial problems: The 15-puzzle.
\newblock \emph{Memory \& cognition}, 33\penalty0 (6):\penalty0 1069--1084, 2005.

\bibitem[Russell \& Norvig(2010)Russell and Norvig]{russell2010artificial}
Stuart~J Russell and Peter Norvig.
\newblock \emph{Artificial intelligence: a modern approach}.
\newblock Pearson Education, 3rd edition, 2010.
\newblock Section 10.1.1: Complexity of Classical Planning.

\bibitem[Shi et~al.(2023)Shi, Lu, and Schwing]{shi2023large}
Zheng Shi, Yujie Lu, and Albert~G Schwing.
\newblock Large language models can be easily distracted by irrelevant context.
\newblock \emph{arXiv preprint arXiv:2302.00093}, 2023.

\bibitem[Shinn et~al.(2023)Shinn, Labash, Gopinath, Lee, Park, Davidson, Zhou, Liang, and Chi]{shinn2023reflexion}
Noah Shinn, Beck Labash, Ashwin Gopinath, Rohan Lee, Maarten Park, Eran Davidson, Denny Zhou, Quoc V~Le Liang, and Ed~Chi.
\newblock Reflexion: Language agents with verbal reinforcement learning.
\newblock \emph{arXiv preprint arXiv:2303.11366}, 2023.

\bibitem[Tomasello(2010)]{tomasello2010origins}
Michael Tomasello.
\newblock \emph{Origins of human communication}.
\newblock MIT press, 2010.

\bibitem[Tversky(2011)]{tversky2011visualizing}
Barbara Tversky.
\newblock Visualizing thought.
\newblock \emph{Topics in cognitive science}, 3\penalty0 (1):\penalty0 159--185, 2011.

\bibitem[Vallati(2015)]{vallati2015automated}
Mauro Vallati.
\newblock Automated synthesis of tetris domains.
\newblock In \emph{Twenty-Ninth AAAI Conference on Artificial Intelligence}, 2015.
\newblock Relevant sentence: "This paper presents an automated approach to synthesizing planning domain models for Tetris puzzles.".

\bibitem[Valmeekam et~al.(2023{\natexlab{a}})Valmeekam, Marquez, Olmo, Sreedharan, and Kambhampati]{valmeekam2023planbench}
Karthik Valmeekam, Matthew Marquez, Alberto Olmo, Sarath Sreedharan, and Subbarao Kambhampati.
\newblock Planbench: An extensible benchmark for evaluating large language models on planning and reasoning about change.
\newblock \emph{Advances in Neural Information Processing Systems}, 36:\penalty0 38975--38987, 2023{\natexlab{a}}.

\bibitem[Valmeekam et~al.(2023{\natexlab{b}})Valmeekam, Nguyen, Le, Li, Kushman, and Polozov]{valmeekam2023planning}
Karthik Valmeekam, Raymond Nguyen, Quoc~V Le, Lin Li, Nate Kushman, and Oleksandr Polozov.
\newblock Planning with large language models for code generation.
\newblock \emph{arXiv preprint arXiv:2302.04761}, 2023{\natexlab{b}}.

\bibitem[Valmeekam et~al.(2024)Valmeekam, Stechly, Gundawar, and Kambhampati]{valmeekam2024planning}
Karthik Valmeekam, Kaya Stechly, Atharva Gundawar, and Subbarao Kambhampati.
\newblock Planning in strawberry fields: Evaluating and improving the planning and scheduling capabilities of lrm o1.
\newblock \emph{arXiv preprint arXiv:2410.02162}, 2024.

\bibitem[Wang et~al.(2023)Wang, Xie, Jiang, Mandlekar, Xiao, Zhu, Fan, and Anandkumar]{wang2023voyager}
Guanzhi Wang, Yuqi Xie, Yunfan Jiang, Ajay Mandlekar, Chaowei Xiao, Yuke Zhu, Linxi Fan, and Anima Anandkumar.
\newblock Voyager: An open-ended embodied agent with large language models.
\newblock \emph{arXiv preprint arXiv:2305.16291}, 2023.

\bibitem[Wang et~al.(2024)Wang, Li, Liu, Wang, and Wang]{wang2024whiteboard}
Pan Wang, Yuchen Li, Jiaqi Liu, Xiaojun Wang, and William~Yang Wang.
\newblock Whiteboard of thought: A framework for llms to plan and reason with visual reminders.
\newblock \emph{arXiv preprint arXiv:2405.19448}, 2024.

\bibitem[Wang et~al.(2022)Wang, Wei, Schuurmans, Le, Chi, Zhou, et~al.]{wang2022self}
Xuezhi Wang, Jason Wei, Dale Schuurmans, Quoc~V Le, Ed~Chi, Denny Zhou, et~al.
\newblock Self-consistency improves chain of thought reasoning in language models.
\newblock \emph{arXiv preprint arXiv:2203.11171}, 2022.

\bibitem[Wei et~al.(2022)Wei, Zhou, Le, and Zhou]{wei2022chain}
Jason Wei, Denny Zhou, Quoc~V Le, and Qu~Zhou.
\newblock Chain-of-thought prompting elicits reasoning in large language models.
\newblock In \emph{Advances in Neural Information Processing Systems}, volume~35, pp.\  24824--24840, 2022.

\bibitem[Wikipedia(2025)]{combinatorial}
Wikipedia.
\newblock Combinatorial optimization, 2025.
\newblock URL \url{https://en.wikipedia.org/wiki/Combinatorial_optimization}.

\bibitem[Wu et~al.(2024)Wu, Jiang, Wang, Zhao, Mandlekar, Anandkumar, Fan, and Zhu]{wu2024imagine}
Junyi Wu, Yunfan Jiang, Guanzhi Wang, Yuheng Zhao, Ajay Mandlekar, Anima Anandkumar, Linxi Fan, and Yuke Zhu.
\newblock Imagine while reasoning in space: Multimodal visualization-of-thought.
\newblock 2024.
\newblock URL \url{https://arxiv.org/abs/2501.07542}.

\bibitem[Yao et~al.(2022)Yao, Zhao, Geng, Lazaridou, Grosse, Cao, Langford, Zhou, Daume~III, et~al.]{yao2022react}
Shunyu Yao, Yuhao Zhao, Yi~Geng, Angeliki Lazaridou, Karthik Grosse, Yuan Cao, John Langford, Denny Zhou, Hal Daume~III, et~al.
\newblock React: Synergizing reasoning and acting in language models.
\newblock In \emph{Advances in Neural Information Processing Systems}, 2022.

\bibitem[Yao et~al.(2023)Yao, Yu, Zhao, et~al.]{yao2023tree}
Shunyu Yao, Dian Yu, Jeffrey Zhao, et~al.
\newblock Tree of thought: Deliberate problem solving with large language models.
\newblock In \emph{Advances in Neural Information Processing Systems}, volume~36, pp.\  478--494, 2023.

\bibitem[Zhong et~al.(2023)Zhong, Ding, Gheini, Bosnic, Dohan, Sag, Zhang, Wang, Yilmaz, et~al.]{zhong2023gpqa}
Yuning Zhong, Ruixiang Ding, Omid Gheini, Predrag Bosnic, David Dohan, Nadav Sag, Ben Zhang, Lin Wang, Yizhong Yilmaz, et~al.
\newblock Gpqa: A benchmark for evaluating general-purpose question answering capabilities of large language models.
\newblock \emph{arXiv preprint arXiv:2305.12479}, 2023.

\end{thebibliography}
\bibliographystyle{iclr2026_conference}

\appendix
\section*{Appendix}

\section{Domain Definitions}
\newcommand{\textsb}[1]{{\fontseries{sb}\selectfont #1}}
\label{sec:appendix_domains}

This section provides formal definitions for the five planning domains introduced in our benchmark. For each domain, we describe the main objective and provide a detailed specification of the available actions, including their purpose, preconditions, and effects.

\subsection{Barman Domain}
The Barman domain models the task of a bartender preparing cocktails. The agent must use two hands to manipulate containers (shots, shakers), ingredients from dispensers, and mix them to create specific cocktails. The state of each object includes its location (on table or held), contents, cleanliness, and for shakers, fill level and whether it has been shaken.

\subsubsection*{Actions}
\begin{enumerate}
    \item \textsb{\texttt{grasp(hand, container)}}: An empty hand picks up a container from the table.
    \begin{itemize}
        \item \textbf{Preconditions}: The container is on the table; the hand is empty.
        \item \textsb{Effects}: The container is no longer on the table; the hand now holds the container and is no longer empty.
    \end{itemize}

    \item \textsb{\texttt{leave(hand, container)}}: A hand places a held container onto the table.
    \begin{itemize}
        \item \textbf{Preconditions}: The hand is holding the container.
        \item \textsb{Effects}: The container is now on the table; the hand becomes empty.
    \end{itemize}

    \item \textsb{\texttt{fill-shot(shot, ingredient, hand1, hand2, dispenser)}}: Fills a clean, empty shot with an ingredient.
    \begin{itemize}
        \item \textbf{Preconditions}: \texttt{hand1} holds the \texttt{shot}; \texttt{hand2} is empty; the \texttt{dispenser} provides the \texttt{ingredient}; the \texttt{shot} is empty and clean.
        \item \textsb{Effects}: The \texttt{shot} now contains the \texttt{ingredient} and is no longer empty or clean (it becomes used).
    \end{itemize}

    \item \textsb{\texttt{refill-shot(shot, ingredient, hand1, hand2, dispenser)}}: Refills a used shot with the same ingredient it previously held.
    \begin{itemize}
        \item \textbf{Preconditions}: \texttt{hand1} holds the \texttt{shot}; \texttt{hand2} is empty; the \texttt{dispenser} provides the \texttt{ingredient}; the \texttt{shot} is empty and was previously used with this ingredient.
        \item \textsb{Effects}: The \texttt{shot} now contains the \texttt{ingredient} and is no longer empty.
    \end{itemize}

    \item \textsb{\texttt{empty-shot(hand, shot, beverage)}}: Empties the contents of a shot.
    \begin{itemize}
        \item \textbf{Preconditions}: The \texttt{hand} is holding the \texttt{shot}; the \texttt{shot} contains the \texttt{beverage}.
        \item \textsb{Effects}: The \texttt{shot} becomes empty.
    \end{itemize}

    \item \textsb{\texttt{clean-shot(shot, beverage, hand1, hand2)}}: Cleans a used, empty shot.
    \begin{itemize}
        \item \textbf{Preconditions}: \texttt{hand1} holds the \texttt{shot}; \texttt{hand2} is empty; the \texttt{shot} is empty and was previously used with the \texttt{beverage}.
        \item \textsb{Effects}: The \texttt{shot} becomes clean and is no longer considered used.
    \end{itemize}

    \item \textsb{\texttt{pour-shot-to-clean-shaker(shot, ingredient, shaker, hand, level-prev, level-next)}}: Pours an ingredient from a shot into a clean, empty shaker.
    \begin{itemize}
        \item \textbf{Preconditions}: The \texttt{hand} holds the \texttt{shot} containing the \texttt{ingredient}; the \texttt{shaker} is empty and clean; the shaker is at \texttt{level-prev}.
        \item \textsb{Effects}: The \texttt{shot} becomes empty; the \texttt{shaker} now contains the \texttt{ingredient}, is no longer empty or clean, becomes unshaken, and its fill level increases to \texttt{level-next}.
    \end{itemize}

    \item \textsb{\texttt{pour-shot-to-used-shaker(shot, ingredient, shaker, hand, level-prev, level-next)}}: Adds a second ingredient to an unshaken shaker.
    \begin{itemize}
        \item \textbf{Preconditions}: The \texttt{hand} holds the \texttt{shot} containing the \texttt{ingredient}; the \texttt{shaker} is unshaken and contains one ingredient; the shaker is at \texttt{level-prev}.
        \item \textsb{Effects}: The \texttt{shot} becomes empty; the \texttt{shaker} now contains the additional ingredient; the shaker's fill level increases to \texttt{level-next}.
    \end{itemize}
    
    \item \textsb{\texttt{empty-shaker(hand, shaker, cocktail, level-prev, level-next)}}: Empties a shaken cocktail from the shaker.
    \begin{itemize}
        \item \textbf{Preconditions}: The \texttt{hand} holds the \texttt{shaker}; the \texttt{shaker} contains a shaken \texttt{cocktail}; the shaker is at \texttt{level-prev}.
        \item \textsb{Effects}: The \texttt{shaker} becomes empty and unshaken; its fill level resets to \texttt{level-next} (empty).
    \end{itemize}
    
    \item \textsb{\texttt{clean-shaker(hand1, hand2, shaker)}}: Cleans an empty shaker.
    \begin{itemize}
        \item \textbf{Preconditions}: \texttt{hand1} holds the \texttt{shaker}; \texttt{hand2} is empty; the \texttt{shaker} is empty.
        \item \textsb{Effects}: The \texttt{shaker} becomes clean.
    \end{itemize}
    
    \item \textsb{\texttt{shake(cocktail, ing1, ing2, shaker, hand1, hand2)}}: Mixes two ingredients in a shaker to create a cocktail.
    \begin{itemize}
        \item \textbf{Preconditions}: \texttt{hand1} holds the \texttt{shaker}; \texttt{hand2} is empty; the \texttt{shaker} contains exactly \texttt{ing1} and \texttt{ing2}; the \texttt{shaker} is unshaken.
        \item \textsb{Effects}: The \texttt{shaker} becomes shaken; it now contains the resulting \texttt{cocktail} instead of the separate ingredients.
    \end{itemize}
    
    \item \textsb{\texttt{pour-shaker-to-shot(cocktail, shot, hand, shaker, level-prev, level-next)}}: Serves a shaken cocktail from a shaker into a shot.
    \begin{itemize}
        \item \textbf{Preconditions}: The \texttt{hand} holds the \texttt{shaker} containing the shaken \texttt{cocktail}; the \texttt{shot} is empty and clean; the shaker is at \texttt{level-prev}.
        \item \textsb{Effects}: The \texttt{shot} now contains the \texttt{cocktail} and is no longer empty or clean; the shaker's fill level decreases to \texttt{level-next}.
    \end{itemize}
\end{enumerate}

\subsection{Elevator Domain}
The Elevator domain involves operating a set of elevators (fast and slow) to transport passengers between floors in a building. Each elevator has a specific capacity and can only access a defined set of floors. The goal is to move all passengers from their origin floors to their destination floors.

\subsubsection*{Actions}
\begin{enumerate}
    \item \textsb{\texttt{move-up-slow(elevator, floor-from, floor-to)}}: Moves a slow elevator up.
    \begin{itemize}
        \item \textbf{Preconditions}: The \texttt{elevator} is at \texttt{floor-from}; \texttt{floor-to} is above \texttt{floor-from}; the \texttt{elevator} can reach \texttt{floor-to}.
        \item \textsb{Effects}: The \texttt{elevator} is now at \texttt{floor-to}.
    \end{itemize}

    \item \textsb{\texttt{move-down-slow(elevator, floor-from, floor-to)}}: Moves a slow elevator down.
    \begin{itemize}
        \item \textbf{Preconditions}: The \texttt{elevator} is at \texttt{floor-from}; \texttt{floor-to} is below \texttt{floor-from}; the \texttt{elevator} can reach \texttt{floor-to}.
        \item \textsb{Effects}: The \texttt{elevator} is now at \texttt{floor-to}.
    \end{itemize}

    \item \textsb{\texttt{move-up-fast(elevator, floor-from, floor-to)}}: Moves a fast elevator up.
    \begin{itemize}
        \item \textbf{Preconditions}: The \texttt{elevator} is at \texttt{floor-from}; \texttt{floor-to} is above \texttt{floor-from}; the \texttt{elevator} can reach \texttt{floor-to}.
        \item \textsb{Effects}: The \texttt{elevator} is now at \texttt{floor-to}.
    \end{itemize}

    \item \textsb{\texttt{move-down-fast(elevator, floor-from, floor-to)}}: Moves a fast elevator down.
    \begin{itemize}
        \item \textbf{Preconditions}: The \texttt{elevator} is at \texttt{floor-from}; \texttt{floor-to} is below \texttt{floor-from}; the \texttt{elevator} can reach \texttt{floor-to}.
        \item \textsb{Effects}: The \texttt{elevator} is now at \texttt{floor-to}.
    \end{itemize}

    \item \textsb{\texttt{board(passenger, elevator, floor, count-prev, count-next)}}: A passenger boards an elevator.
    \begin{itemize}
        \item \textbf{Preconditions}: The \texttt{passenger} and \texttt{elevator} are at the same \texttt{floor}; the elevator's passenger count is \texttt{count-prev}; the elevator has capacity for another passenger.
        \item \textsb{Effects}: The \texttt{passenger} is now on board the \texttt{elevator}; the elevator's passenger count becomes \texttt{count-next}.
    \end{itemize}

    \item \textsb{\texttt{leave(passenger, elevator, floor, count-prev, count-next)}}: A passenger leaves an elevator.
    \begin{itemize}
        \item \textbf{Preconditions}: The \texttt{passenger} is on board the \texttt{elevator}; the \texttt{elevator} is at the specified \texttt{floor}; the elevator's passenger count is \texttt{count-prev}.
        \item \textsb{Effects}: The \texttt{passenger} is now at the \texttt{floor}; the elevator's passenger count becomes \texttt{count-next}.
    \end{itemize}
\end{enumerate}

\subsection{Parking Domain}
The Parking domain involves rearranging cars parked at curbs. Each curb can hold at most two cars: one parked at the curb and one double-parked behind it. A car cannot move if another car is parked behind it.

\subsubsection*{Key Predicates}
\begin{itemize}
    \item \textsb{\texttt{clear(car)}}: True if no car is double-parked behind this car.
    \item \textsb{\texttt{clear(curb)}}: True if the curb is empty.
\end{itemize}

\subsubsection*{Actions}
\begin{enumerate}
    \item \textsb{\texttt{move-curb-to-curb(car, curb-from, curb-to)}}: A single-parked car moves to an empty curb.
    \begin{itemize}
        \item \textbf{Preconditions}: \texttt{car} is at \texttt{curb-from}; \texttt{car} is clear; \texttt{curb-to} is clear.
        \item \textsb{Effects}: \texttt{curb-from} becomes clear; \texttt{car} is now at \texttt{curb-to}, which is no longer clear.
    \end{itemize}

    \item \textsb{\texttt{move-curb-to-car(car-move, curb-from, car-ahead)}}: A single-parked car double-parks behind another car.
    \begin{itemize}
        \item \textbf{Preconditions}: \texttt{car-move} is at \texttt{curb-from}; \texttt{car-move} is clear; \texttt{car-ahead} is clear.
        \item \textsb{Effects}: \texttt{curb-from} becomes clear; \texttt{car-move} is now behind \texttt{car-ahead}; \texttt{car-ahead} is no longer clear.
    \end{itemize}

    \item \textsb{\texttt{move-car-to-curb(car-move, car-ahead, curb-to)}}: A double-parked car moves to an empty curb.
    \begin{itemize}
        \item \textbf{Preconditions}: \texttt{car-move} is behind \texttt{car-ahead}; \texttt{car-move} is clear; \texttt{curb-to} is clear.
        \item \textsb{Effects}: \texttt{car-ahead} becomes clear; \texttt{car-move} is now at \texttt{curb-to}, which is no longer clear.
    \end{itemize}

    \item \textsb{\texttt{move-car-to-car(car-move, car-from, car-to)}}: A double-parked car moves to double-park behind a different car.
    \begin{itemize}
        \item \textbf{Preconditions}: \texttt{car-move} is behind \texttt{car-from}; \texttt{car-move} is clear; \texttt{car-to} is clear.
        \item \textsb{Effects}: \texttt{car-from} becomes clear; \texttt{car-move} is now behind \texttt{car-to}; \texttt{car-to} is no longer clear.
    \end{itemize}
\end{enumerate}

\subsection{Tetris Domain}
The Tetris domain involves moving and rotating Tetris pieces on a grid. Pieces can be one-square, two-square straight, or three-square L-shaped. A piece can only move or rotate into adjacent positions that are clear (empty).

\subsubsection*{Actions}
\begin{enumerate}
    \item \textsb{\texttt{move\_square(pos-from, pos-to, piece)}}: Moves a one-square piece.
    \begin{itemize}
        \item \textbf{Preconditions}: \texttt{piece} occupies \texttt{pos-from}; \texttt{pos-to} is clear; \texttt{pos-from} and \texttt{pos-to} are adjacent.
        \item \textsb{Effects}: \texttt{pos-from} becomes clear; \texttt{pos-to} is now occupied by \texttt{piece}.
    \end{itemize}

    \item \textsb{\texttt{move\_two(pos-old, pos-pivot, pos-new, piece)}}: Moves or rotates a two-square piece.
    \begin{itemize}
        \item \textbf{Preconditions}: \texttt{piece} occupies \texttt{pos-old} and \texttt{pos-pivot}; \texttt{pos-new} is clear; \texttt{pos-pivot} and \texttt{pos-new} are adjacent.
        \item \textsb{Effects}: \texttt{pos-old} becomes clear; \texttt{pos-new} is now occupied by \texttt{piece}.
    \end{itemize}

    \item \textsb{\texttt{move\_l\_right(pA, pB, pC, pD, pE, pMid, piece)}}: Moves or rotates an L-piece right.
    \begin{itemize}
        \item \textbf{Preconditions}: \texttt{piece} occupies \texttt{pA, pB, pC}; \texttt{pD, pE} are clear; positions are correctly adjacent for a rightward move.
        \item \textsb{Effects}: \texttt{pA, pB} become clear; \texttt{pD, pE} are now occupied by \texttt{piece}.
    \end{itemize}
    
    \item \textsb{\texttt{move\_l\_left(pA, pB, pC, pD, pE, piece)}}: Moves or rotates an L-piece left.
    \begin{itemize}
        \item \textbf{Preconditions}: \texttt{piece} occupies \texttt{pA, pB, pC}; \texttt{pD, pE} are clear; positions are correctly adjacent for a leftward move.
        \item \textsb{Effects}: \texttt{pA, pC} become clear; \texttt{pD, pE} are now occupied by \texttt{piece}.
    \end{itemize}
    
    \item \textsb{\texttt{move\_l\_up(pA, pB, pC, pD, pE, pMid, piece)}}: Moves or rotates an L-piece up.
    \begin{itemize}
        \item \textbf{Preconditions}: \texttt{piece} occupies \texttt{pA, pB, pC}; \texttt{pD, pE} are clear; positions are correctly adjacent for an upward move.
        \item \textsb{Effects}: \texttt{pB, pC} become clear; \texttt{pD, pE} are now occupied by \texttt{piece}.
    \end{itemize}
    
    \item \textsb{\texttt{move\_l\_down(pA, pB, pC, pD, pE, piece)}}: Moves or rotates an L-piece down.
    \begin{itemize}
        \item \textbf{Preconditions}: \texttt{piece} occupies \texttt{pA, pB, pC}; \texttt{pD, pE} are clear; positions are correctly adjacent for a downward move.
        \item \textsb{Effects}: \texttt{pA, pC} become clear; \texttt{pD, pE} are now occupied by \texttt{piece}.
    \end{itemize}
\end{enumerate}

\subsection{Floor Tiles Domain}
The Floor Tiles domain involves robots painting a grid of tiles. Each robot can hold one color at a time and moves between adjacent tiles. A robot can paint an adjacent tile (above or below) with its current color, provided the tile is clear. Once a tile is painted, it cannot be occupied.

\subsubsection*{Actions}
\begin{enumerate}
    \item \textsb{\texttt{change-color(robot, color-from, color-to)}}: A robot changes its held paint color.
    \begin{itemize}
        \item \textbf{Preconditions}: The \texttt{robot} is holding \texttt{color-from}; \texttt{color-to} is an available color.
        \item \textsb{Effects}: The \texttt{robot} is now holding \texttt{color-to}.
    \end{itemize}

    \item \textsb{\texttt{paint-up(robot, tile-paint, tile-robot, color)}}: A robot paints the tile above its current position.
    \begin{itemize}
        \item \textbf{Preconditions}: \texttt{robot} is at \texttt{tile-robot}; \texttt{tile-paint} is directly above \texttt{tile-robot}; \texttt{tile-paint} is clear; \texttt{robot} is holding \texttt{color}.
        \item \textsb{Effects}: \texttt{tile-paint} is now painted with \texttt{color} and is no longer clear.
    \end{itemize}

    \item \textsb{\texttt{paint-down(robot, tile-paint, tile-robot, color)}}: A robot paints the tile below its current position.
    \begin{itemize}
        \item \textbf{Preconditions}: \texttt{robot} is at \texttt{tile-robot}; \texttt{tile-paint} is directly below \texttt{tile-robot}; \texttt{tile-paint} is clear; \texttt{robot} is holding \texttt{color}.
        \item \textsb{Effects}: \texttt{tile-paint} is now painted with \texttt{color} and is no longer clear.
    \end{itemize}

    \item \textsb{\texttt{up(robot, tile-from, tile-to)}}: A robot moves one tile up.
    \begin{itemize}
        \item \textbf{Preconditions}: \texttt{robot} is at \texttt{tile-from}; \texttt{tile-to} is directly above \texttt{tile-from}; \texttt{tile-to} is clear.
        \item \textsb{Effects}: \texttt{robot} is now at \texttt{tile-to}; \texttt{tile-from} becomes clear; \texttt{tile-to} is no longer clear.
    \end{itemize}

    \item \textsb{\texttt{down(robot, tile-from, tile-to)}}: A robot moves one tile down.
    \begin{itemize}
        \item \textbf{Preconditions}: \texttt{robot} is at \texttt{tile-from}; \texttt{tile-to} is directly below \texttt{tile-from}; \texttt{tile-to} is clear.
        \item \textsb{Effects}: \texttt{robot} is now at \texttt{tile-to}; \texttt{tile-from} becomes clear; \texttt{tile-to} is no longer clear.
    \end{itemize}

    \item \textsb{\texttt{right(robot, tile-from, tile-to)}}: A robot moves one tile right.
    \begin{itemize}
        \item \textbf{Preconditions}: \texttt{robot} is at \texttt{tile-from}; \texttt{tile-to} is directly to the right of \texttt{tile-from}; \texttt{tile-to} is clear.
        \item \textsb{Effects}: \texttt{robot} is now at \texttt{tile-to}; \texttt{tile-from} becomes clear; \texttt{tile-to} is no longer clear.
    \end{itemize}

    \item \textsb{\texttt{left(robot, tile-from, tile-to)}}: A robot moves one tile left.
    \begin{itemize}
        \item \textbf{Preconditions}: \texttt{robot} is at \texttt{tile-from}; \texttt{tile-to} is directly to the left of \texttt{tile-from}; \texttt{tile-to} is clear.
        \item \textsb{Effects}: \texttt{robot} is now at \texttt{tile-to}; \texttt{tile-from} becomes clear; \texttt{tile-to} is no longer clear.
    \end{itemize}
\end{enumerate}

\section{Overview of Prompts}
\label{sec:appendix_prompts}

\renewcommand{\arraystretch}{1.12}

In this section, we provide a mapping between the main inference functions and their associated prompts, clarifying how the model is prompted at each stage of the search and diagrammatic reasoning pipeline. For each function, we specify: (1) the function name, (2) its core functionality, (3) where it is called in the search process, (4) the key points of its prompt, and (5) its input parameters. This overview enables readers to understand how the model is guided through each step of multimodal reasoning.

\bigskip

\setlength\LTleft{0pt}
\setlength\LTright{0pt}

\begin{longtable}{|>{\raggedright\arraybackslash\ttfamily\small}p{1.5cm}|>{\raggedright\arraybackslash}p{2cm}|>{\raggedright\arraybackslash}p{2.5cm}|>{\raggedright\arraybackslash}p{3cm}|>{\raggedright\arraybackslash}p{3cm}|}
\hline
\textbf{Function Name} & \textbf{Functionality} & \textbf{Where Called} & \textbf{Key Prompt Points} & \textbf{Input Parameters (Description)} \\
\hline
\endfirsthead
\hline
\textbf{Function Name} & \textbf{Functionality} & \textbf{Where Called} & \textbf{Key Prompt Points} & \textbf{Input Parameters (Description)} \\
\hline
\endhead

next\newline\_action &
Selects next best action, generates reasoning, action, and new state description. &
During child state generation (generate\_child\_states). &
- Presents problem, initial, current, and goal state (text and images)\newline
- Lists possible actions\newline
- Requests reasoning, action, and new state in code blocks\newline
- Handles uniqueness and previous errors if present &
- Problem description (text)\newline
- Possible actions (list)\newline
- Initial, current, goal state objects (text, diagrams, schemas)\newline
- Model name, temperature\newline
- Chosen actions\newline
- Previous attempt/error (optional) \\
\hline

generate\newline\_diagram\newline\_schema &
Generates diagram schema for a child state after an action. &
During child state diagram generation (generate\_diagrams). &
- Presents problem, initial, current, and new state (text, diagrams, schemas)\newline
- Requests one statement per object (position, size, status, identifier)\newline
- Enforces object count consistency\newline
- Handles previous attempt/error if present &
- Problem description (text)\newline
- Initial state object (with schema)\newline
- Child state object (with action, parent, state description)\newline
- Model name, temperature\newline
- Previous attempt/error (optional) \\
\hline

test\newline\_diagram\newline\_schema &
Verifies correctness of a generated diagram schema for a child state. &
After schema generation, before code generation (generate\_diagrams). &
- Presents problem, initial, current, and child state (text, diagrams, schemas)\newline
- Validates object count, affected object status, and consistency\newline
- Requires yes/no answer and error summary &
- Problem description (text)\newline
- Initial state object (with schema)\newline
- Child state object (with schema, action, parent)\newline
- Model name \\
\hline

generate\newline\_diagram\newline\_code &
Generates Matplotlib code to visualize a child state based on its schema. &
After schema validation (generate\_diagrams). &
- Presents problem, initial/current/child state (text, diagrams, schemas)\newline
- Provides example code and reasoning if available\newline
- Requests code to visualize all objects as described in schema\newline
- Enforces clarity, no overlaps, correct labeling, plausibility\newline
- Handles previous attempt/error if present &
- Domain name\newline
- Problem description (text)\newline
- Child state object (with schema, action, parent)\newline
- Initial state object (with code, diagram)\newline
- Model name, temperature\newline
- Save path for image\newline
- Previous attempt/error (optional) \\
\hline

test\newline\_diagram &
Verifies correctness and clarity of a generated diagram image for a child state. &
After code execution, before accepting diagram (generate\_diagrams). &
- Presents problem, current/child state (text, diagrams, schemas)\newline
- Validates object presence, status, labeling, readability, plausibility\newline
- Requires yes/no answer and error summary &
- Problem description (text)\newline
- Child state object (with schema, diagram, parent)\newline
- Domain name\newline
- Model name \\
\hline

check\newline\_action\newline\_validity &
Checks if a proposed action and resulting state are valid (local verification). &
After diagram generation, before path/global verification (generate\_child\_states). &
- Presents problem, initial/current/goal state (text, diagrams, schemas)\newline
- Lists action path, action taken, new state description, and diagram\newline
- Checks preconditions, effects, diagram accuracy, reasoning\newline
- Requires yes/no answer and error summary &
- Problem description (text)\newline
- Current state object (with schema, diagram)\newline
- Action taken (string)\newline
- New state description (text)\newline
- New state object (with schema, diagram)\newline
- Goal state object\newline
- Possible actions (list)\newline
- Initial state object\newline
- Model name \\
\hline

check\newline\_action\newline\_path &
Checks if the entire action path from initial to current state is valid (global verification). &
After local verification, before accepting child state (generate\_child\_states). &
- Presents problem, initial/current/goal state (text, diagrams, schemas)\newline
- Lists full action path and parent state\newline
- Checks preconditions, effects, diagram accuracy for the whole path\newline
- Requires yes/no answer and reasoning &
- Problem description (text)\newline
- Initial state object (with schema, diagram)\newline
- Current state object (with schema, diagram, parent)\newline
- Goal state object\newline
- Actions (list)\newline
- Possible actions (list)\newline
- Model name \\
\hline

rank\newline\_states &
Ranks candidate states at a given depth by proximity to the goal state (for beam search). &
After child state generation, before pruning (beam\_search). &
- Presents problem, goal state, and all candidate states (text, diagrams, action paths)\newline
- Requests ranking based on number of goal constraints satisfied\newline
- Requires ranking code block and reasoning &
- List of state objects (with id, description, diagram, action path)\newline
- Problem description (text)\newline
- Goal state object\newline
- Model name \\
\hline

ini\_g\newline\_diagrams &
Generates and verifies diagrams for initial and goal states. &
At the start of search, during problem setup (setup). &
- Presents problem, initial/goal state (text)\newline
- Requests diagram schema and code for initial/goal state\newline
- Enforces object coverage, status, and visual consistency\newline
- Verifies schema and diagram correctness\newline
- Handles retries and error feedback &
- Problem name, domain name\newline
- Problem description (text)\newline
- Initial state object (with text)\newline
- Goal state object (with text)\newline
- Model name, temperature\newline
- Parameters (e.g., max attempts)\newline
- Error message, previous attempt (optional) \\
\hline

generate\newline\_diagram\newline\_schema\newline\_ini\_g &
Generates diagram schema for initial or goal state. &
Within ini\_g\_diagrams (problem setup). &
- Presents problem, initial and goal state descriptions\newline
- Requests one statement per object (position, size, status, identifier)\newline
- Handles previous attempt/error if present &
- Problem description (text)\newline
- Initial state object (with text)\newline
- Goal state object (with text)\newline
- Model name, temperature\newline
- Domain name\newline
- Goal flag (bool)\newline
- Previous attempt/error (optional) \\
\hline

test\newline\_diagram\newline\_schema\newline\_ini\_g &
Verifies correctness of initial/goal state diagram schema. &
Within ini\_g\_diagrams (after schema generation). &
- Presents problem, state description, and schema\newline
- Validates object coverage and status\newline
- Requires yes/no answer and error summary &
- Problem description (text)\newline
- Initial state object (with schema)\newline
- Goal state object (with schema)\newline
- Model name\newline
- Goal flag (bool) \\
\hline

generate\newline\_diagram\newline\_code\newline\_ini\_g &
Generates Matplotlib code for initial or goal state visualization. &
Within ini\_g\_diagrams (after schema validation). &
- Presents problem, state description, and schema\newline
- Provides code and reasoning\newline
- Requests code to visualize all objects as described in schema\newline
- Enforces clarity, labeling, plausibility\newline
- Handles previous attempt/error if present &
- Domain name\newline
- Problem description (text)\newline
- Initial state object (with schema)\newline
- Goal state object (with schema)\newline
- Model name, temperature\newline
- Save path for image\newline
- Goal flag (bool)\newline
- Previous attempt/error (optional) \\
\hline

test\newline\_diagram\newline\_ini\_g &
Verifies correctness and clarity of initial/goal state diagram image. &
Within ini\_g\_diagrams (after code execution). &
- Presents problem, state description, and diagram image\newline
- Validates object presence, status, labeling, consistency, plausibility\newline
- Requires yes/no answer and error summary &
- Problem description (text)\newline
- Initial state object (with schema, diagram)\newline
- Goal state object (with schema, diagram)\newline
- Domain name\newline
- Model name\newline
- Goal flag (bool) \\
\hline

is\_unique\newline\_action &
Checks if a proposed action leads to a unique child state from the current state. &
During child state generation in generate\_child\_states. &
- Presents problem, current state, action taken, and new state description\newline
- Lists previously explored actions and resulting states\newline
- Asks if the new action/state is unique\newline
- Requires yes/no answer and explanation &
- Problem description (text)\newline
- Current state object (with child states)\newline
- Action taken (string)\newline
- New state description (text)\newline
- Model name \\
\hline

check\newline\_goal\newline\_state &
Checks if the current state satisfies all goal state constraints. &
At each node expansion in beam\_search. &
- Presents problem, initial, current, and goal state (text, diagrams)\newline
- Lists action path\newline
- Asks if current state matches all goal constraints\newline
- Requires yes/no answer and step-by-step reasoning &
- Problem description (text)\newline
- Current state object (with diagram)\newline
- Goal state object (with diagram)\newline
- Initial state object (with diagram)\newline
- Model name \\
\hline

\end{longtable}

This mapping clarifies the modular structure of the inference pipeline and the precise role of each prompt in guiding the model's multimodal reasoning. Each function is responsible for a distinct aspect of the search or verification process, and the prompts are carefully designed to enforce correctness, clarity, and consistency at every stage. The table above can be used as a reference for understanding or extending the codebase, as well as for reproducing or adapting the prompting strategy to new domains or tasks.

\section{Code Structure and Organization}
\label{sec:appendix_code}

The codebase is organized to support multimodal planning and diagrammatic reasoning across multiple domains. Each domain (e.g., \texttt{blocksworld}, \texttt{barman}, \texttt{elevator}, \texttt{parking}, \texttt{tetris}, \texttt{tiles}) is self-contained, with a consistent folder structure and supporting scripts. Below, we describe the main components and their roles.

\subsection{Top-Level Structure}

\begin{itemize}
    \item \textbf{Domain Folders:} Each domain (e.g., \texttt{blocksworld}, \texttt{barman}, etc.) is a top-level folder containing all files and subfolders needed to run our method on the instances of that domain.
    \item \textbf{Shared Scripts:} At the root, scripts such as \texttt{search.py}, \texttt{inference.py}, and diagram generation scripts are provided for general use across domains.
    \item \textbf{Utilities:} Files like \texttt{requirements.txt} and \texttt{Readme.md} provide environment setup and documentation.
\end{itemize}

\subsection{Key Scripts and Their Roles}

\begin{itemize}
    \item \textsb{\texttt{search.py}}: The main search and script. Implements the graph of thought algorithm with beam search and backtracking, manages state expansion, diagram generation, and verification. For each problem instance, it creates a subfolder (e.g., \texttt{blocksworld\_instance\_1/}) and stores all intermediate and final results, including state diagrams, code, and logs.
    \item \textsb{\texttt{inference.py}}: Contains all functionalities facilitated by LMMs and prompting logic, including functions for generating and verifying actions, diagram schemas, code, and state rankings. This script is the interface between the search process and the language model.
    \item \textsb{\texttt{visual\_thinking.py}}: Main pipeline script. Orchestrates the batch processing of multiple problem instances for a given domain. For each instance, it sets up the directory structure, runs our diagrammatic search, and validates the resulting plan using VAL. It also manages the output and validation logs.
    \item \textsb{\texttt{initial\_diagram\_schema.py}}, \texttt{initial\_diagram\_code.py}, \texttt{initial\_goal\_diagram\_code.py}: Scripts for generating and storing the first initial and goal conceptual diagrams for the domain and their schemas. Outputs are stored in the \texttt{initial\_conceptual\_diagram/} folder.
    \item \textsb{\texttt{PDDL\_tranlation.py}} (per domain): Handles translation between natural language and PDDL for that domain, including prompt templates and in context examples. 
\end{itemize}

\subsection{Instance and State Folder Structure}

For each problem instance (e.g., \texttt{blocksworld\_instance\_1/}), the following structure is created during search:

\begin{itemize}
    \item \texttt{state\_X/}: For each explored state, a folder is created containing:
    \begin{itemize}
        \item \texttt{diagram.png}: The generated diagram for the state.
        \item \texttt{diagram\_code.py}: The code used to generate the diagram.
        \item \texttt{diagram\_schema.txt}: The schema describing the diagram.
        \item \texttt{info.txt}: Metadata about the state, including parent, action taken, and reasoning.
        \item \texttt{attempts/}: Subfolders for storing all attempts at generating child states and diagrams, including error logs.
    \end{itemize}
    \item \texttt{goal\_state/}: If a goal is found, this folder contains the final state information and a copy of all diagrams along the solution path.
    \item \texttt{ranking/}: Stores state ranking information at each search depth.
    \item \texttt{output.txt}, \texttt{plan.pddl}, \texttt{val\_output.txt}: Output logs, the generated plan, and validation results.
\end{itemize}

\subsection{Domain Examples}

The repository attached in the supplementary material includes several PDDL domains (\texttt{blocksworld}, \texttt{barman}, \texttt{elevator}, \texttt{parking}, \texttt{tetris}, \texttt{tiles}), each with a complete set of PDDL files, instance generators, randomly generated instances, and translations of the domain and a random state and initial conceptual diagrams generated using pipeline. For \texttt{blocksworld}, we also provide full examples of solved problems, including all intermediate diagrams and logs, to facilitate reproducibility and intuitive understanding of the pipeline. To view the sequence of diagrams produced for a successfully solved instance, refer to the \texttt{goal\_state} folder within each instance’s directory. This folder contains the chain of diagrams generated by the pipeline from the initial state to the goal state for that instance.

\bigskip

This modular structure allows for easy extension to new domains and facilitates reproducibility, inspection, and further research.

\end{document}